\documentclass{IEEEphot}
\usepackage{cite}

\usepackage{amsmath,amssymb,amsfonts}
\usepackage{algorithmic}
\usepackage{algorithm}
\usepackage{booktabs}
\usepackage{graphicx,color}
\usepackage{multirow}
\usepackage{textcomp}
\usepackage{xcolor}

\usepackage{censor} 
\renewcommand{\xblackout}[1]{\ifanonymous\blackout{#1}\else#1\fi}

\usepackage{hyperref}
\usepackage{cleveref}
\makeatletter
\renewcommand{\p@subsection}{\thesection-}
\renewcommand{\p@subsubsection}{\thesection-\thesubsection}
\makeatother
\hypersetup{pdfpagemode=UseNone}
\usepackage{orcidlink} 
\let\labelindent\relax
\usepackage{enumitem}
\def\BibTeX{{\rm B\kern-.05em{\sc i\kern-.025em b}\kern-.08em
    T\kern-.1667em\lower.7ex\hbox{E}\kern-.125emX}}
\AtBeginDocument{\definecolor{tmlcncolor}{cmyk}{0.93,0.59,0.15,0.02}\definecolor{NavyBlue}{RGB}{0,86,125}}

\usepackage{siunitx}
\providecommand{\unit}{\si}
\providecommand{\qty}{\SI}
\DeclareSIUnit{\byte}{B}
\DeclareSIUnit{\pixel}{px}
\DeclareSIUnit\megapixel{MP}
\DeclareSIUnit{\decibel}{dB}
\DeclareSIUnit{\fps}{fps}
\DeclareSIUnit{\USD}{\text{USD}}
\DeclareSIUnit{\ggrav}{\ensuremath{g_{\text{grav}}}}
\DeclareSIUnit{\Mbps}{Mbps}
\DeclareSIUnit{\Gbps}{Gbps}
\DeclareSIUnit{\ppm}{ppm}
\DeclareSIUnit\points{points}
\DeclareSIUnit{\herz}{Hz}

\hypersetup{
    pdftoolbar=true,        
    pdfmenubar=true,        
    pdffitwindow=false,     
    pdfstartview={FitH},    
    pdftitle={SAFARI: Semantic And Frontier-driven Autonomous Risk Inspection},    
    pdfsubject={IEEE Transactions on Field Robotics},   
    colorlinks=true,
    linkcolor=blue,
    linkbordercolor=white,
    filecolor=magenta,
    citecolor=red,
    urlcolor=cyan
    }

\def\OJlogo{\vspace{-4pt}$<$Society logo(s) and publication title will appear here.$>$}
\def\seclogo{\vspace{10pt}$<$Society logo(s) and publication title will appear here.$>$}

\def\authorrefmark#1{\ensuremath{^{\textbf{#1}}}}

\begin{document}
\receiveddate{XX Month, XXXX}
\reviseddate{XX Month, XXXX}
\accepteddate{XX Month, XXXX}
\publisheddate{XX Month, XXXX}
\currentdate{XX Month, XXXX}
\doiinfo{XXXX.2022.1234567}

\markboth{SAFARI}{\xblackout{Kafkasyali \& Toader {et al.}}}

\title{SAFARI: Semantic- And Frontier-driven Autonomous Risk Inspection}

\newif\ifanonymous
\anonymousfalse 
\ifanonymous\else
\hypersetup{pdfauthor={Ugur Kafkasyali, Mihnea Toader, Julia Richter, Turcan Tuna, Cesar Cadena, Marco Hutter}}
\fi


\DeclareRobustCommand{\anontext}[2]{%
  \ifanonymous
    \textcolor{black}{#2}%
  \else
    #1%
  \fi
}

\author{
\xblackout{Ugur Kafkasyali}$^{\anontext{\orcidlink{0009-0000-5080-7206}}{}}$\authorrefmark{1}$^*$,
\xblackout{Mihnea Toader}$^{\anontext{\orcidlink{0009-0009-4177-3660}}{}}$\authorrefmark{1}$^*$,\\
\xblackout{Julia Richter}$^{\anontext{\orcidlink{0009-0001-5477-6440}}{}}$\authorrefmark{1}
(\xblackout{Graduate Student Member, IEEE}),
\xblackout{Turcan~Tuna}$^{\anontext{\orcidlink{0000-0001-8662-4890}}{}}$\authorrefmark{1}
(\xblackout{Graduate Student Member, IEEE}),
\xblackout{Cesar~Cadena}$^{\anontext{\orcidlink{0000-0002-2972-6011}}{}}$\authorrefmark{1}
(\xblackout{Member, IEEE}), and
\xblackout{Marco~Hutter}$^{\anontext{\orcidlink{0000-0002-4285-4990}}{}}$\authorrefmark{1}
(\xblackout{Member, IEEE})
}
\affil{\xblackout{Robotic Systems Lab, Swiss Federal Institute of Technology
(ETH Zürich), 8092 Zürich, Switzerland}}
\corresp{Corresponding author: \xblackout{Mihnea Toader (email: \texttt{toaderm@ethz.ch})}}
\authornote{
    *Equal contribution\\
    \\
    \xblackout{
    The research leading to these results has received funding from armasuisse Science and Technology and was supported by the Swiss National Science Foundation (SNSF) as part of the project No. 227617.
    }
}
\begin{abstract}
Robotic exploration can reduce human exposure to hazardous environments by autonomously searching unknown areas for objects of interest while constructing a complete, risk-aware map.
However, existing autonomous exploration approaches typically prioritize geometric coverage or information gain, without explicitly coupling semantic object detection and the risk associated with operating near potentially hazardous objects.
Moreover, many systems have been demonstrated primarily in simulation or controlled environments, leaving limited validation of risk-aware semantic exploration in complex, unstructured field settings.
To address this gap, we propose SAFARI, a framework that treats object detection as a native channel of a unified probabilistic map, thereby jointly informing exploration and planning with both geometric and semantic evidence. 
We evaluated SAFARI in 11 missions across three field sites on a quadruped platform. The framework achieved an average coverage of \qty{98.7}{\percent} of the accessible mission area, without operator intervention for \qty{97.6}{\percent} of mission time, and detected objects with characteristic dimensions as small as \SI{15}{\centi\metre}.
Lastly, through ablation studies and a detailed analysis of system behavior, we show that object-size awareness and safety-aware path planning are crucial for effective hazardous-object search.
\end{abstract}



\maketitle

\section{INTRODUCTION}
Autonomous exploration is fundamental for many robotic applications involving reconnaissance, such as disaster response, search and rescue, and forest inventory. A key challenge in these tasks is that the environments may contain unknown hazards. 
Such hazards can pose a threat not only to the robot but also to the environment or the robot operators. To make such tasks as safe as possible, reduction of human interaction through increased robot autonomy is a must~\cite{baudoin_using_2011}. 

Conducting reconnaissance autonomously can reduce human exposure to hazardous environments, while some tasks still require fine motor skills, precise judgment, and other capabilities that are not yet reliably reproducible by robots.
Hence, a robot-led reconnaissance mission prior to human involvement directly addresses this problem: given an area to explore, the robot should detect objects of potential interest (OPIs) -- hazardous items or any other mission-relevant class -- and create a map that marks areas as dangerous, observed to be safe, or unobserved.

\begin{figure*}[t]
    \centering
    \includegraphics[width=1.0\linewidth]{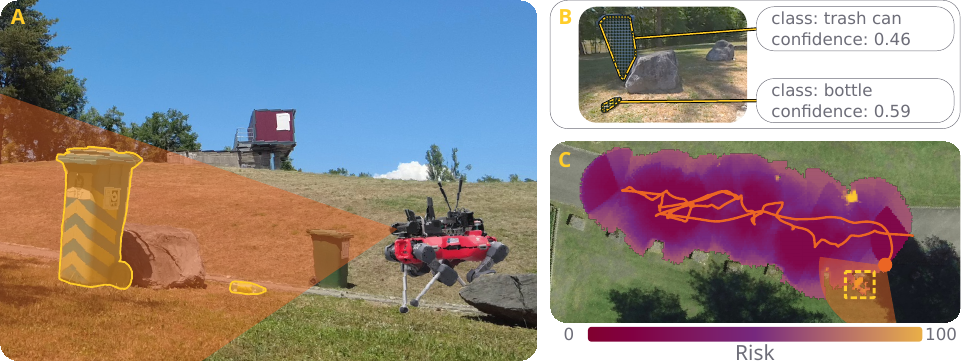}
    \caption{Illustration of the mission setup. Given a bounded region, the robot autonomously explores, while simultaneously detecting and avoiding collision with objects of potential interest. \textbf{(A)} Third-person view of an ANYbotics ANYmal D approaching and correctly detecting two OPIs: a trash can and a littered bottle. \textbf{(B)} First-person view of the robot's perspective, overlaid with segmentation masks and detection output. \textbf{(C)} Accumulated risk layer overlaid on satellite imagery of the explored area. The robot's trajectory is depicted in orange.}
    \label{fig:overview_viz}
\end{figure*}

Autonomous exploration around OPIs is challenging. 
To achieve complete coverage efficiently, the robot must continuously assess traversability, determine which regions are safe with sufficient confidence, and minimize mission duration while accounting for uncertainty in its observations. 
Depending on the task, a missed detection can range from inconvenient to catastrophic.
Additionally, the range at which it is possible to detect visible OPIs decreases with object size: the robot must get closer to small objects to reliably detect them, whereas larger objects are discernible from farther away.
Finally, traversability and OPI placement must be jointly reasoned over when deciding which region to explore next, since considering either in isolation omits critical information.

In this work, we present SAFARI, a detection, mapping, and path planning pipeline for autonomous exploration and reporting of OPIs in unstructured outdoor environments.
SAFARI accumulates depth-supported positive detections and object-size-dependent negative evidence in a persistent risk field. The resulting observation demand guides frontier and viewpoint selection together with geometric traversability and risk.
SAFARI consists of three modules: \textit{1)} a detection module that processes an incoming RGB image, captured from a top-mounted RGB-D camera; \textit{2)} a mapping module that accumulates the depth measurements from the sensor and the semantic data from the detection module to form a probabilistic, persistent representation of the environment; \textit{3)} a path planning module that uses this representation for coverage planning of the area. These modules are detailed in \Cref{sec:observation_generation_draft,sec:mapping_draft,sec:planning_draft}, respectively.
\Cref{fig:overview_viz} illustrates the mission setup and operating principle of the system. 
We verified real-world applicability through outdoor deployment campaigns across three locations consisting of 11 missions, shown in \Cref{sec:experiments}.

Our contributions are as follows:

\begin{itemize}[topsep=0pt, partopsep=0pt]
    \item A probabilistic mapping framework that extends the lambda field formulation \cite{laconte_novel_2021} (see \Cref{sec:preliminaries_draft}) with semantic hazard evidence and uncertainty estimation. An object-size-dependent detectability model weights negative evidence from visible regions without detections, supporting assessment of whether a region is sufficiently inspected under a chosen size hypothesis.

    \item A hierarchical path planner that combines frontier-based exploration with viewpoint optimization and uses the resulting observation demand, together with geometric traversability and risk, to guide further inspection.

    \item Extensive field validation comprising 11 missions across three locations, of which three are primary mission runs with a fixed parameter set and the remainder are ablations to provide detailed insights into the proposed framework.

\end{itemize}

\section{RELATED WORK}
\label{sec:related_work}

Our proposed approach draws from probabilistic mapping, semantically informed planning, and information-gain driven exploration. We begin by reviewing each of these research areas separately and then discuss systems that, like ours, tackle the complete \textit{search and report} task formulation, where we identify important limitations in their scale and architecture.
\subsection{Occupancy and Hazard Mapping}
Traditionally, the center of any autonomous exploration pipeline is its memory representation \cite{kim_evidential_2024, deng_see-csom_2024, matez-bandera_voxeland_2024}.
Typical representations include occupancy grids, which discretize the operating space and update cell states from sensor data, typically via Bayesian log-odds fusion \cite{miles_terrain-aware_2023}; voxel-based maps, which extend this discretization to full 3-D volumes and are often paired with signed-distance fields for continuous surface reconstruction \cite{oleynikova_voxblox_2017}; and topological or viewpoint graphs, which encode the environment as a set of connected nodes rather than a dense spatial grid, enabling more compact, high-level reasoning for planning \cite{liang_hdplanner_2024, dang_graph-based_2020}.
A rich body of work has extended the classical voxel grid to improve robustness and continuity. 
Bayesian Kernel Inference methods propagate sparse measurements across neighboring cells, with recent variants incorporating semantic uncertainty \cite{kim_evidential_2024}, adaptive kernel lengths for sharp object boundaries \cite{deng_see-csom_2024}, and both geometric and semantic evidence \cite{matez-bandera_voxeland_2024}.
A fundamental limitation of the discrete Bayesian occupancy formulation is that the probability of an event is inherently coupled to the size of a voxel. 
Laconte et al. \cite{laconte_novel_2021, laconte_dynamic_2021} introduce the lambda field as an alternative mapping framework, designed for probabilistic mapping of unstructured environments and risk-aware path planning. Rather than representing the occupancy of a cell with a probability value, the spatial intensity of an arbitrary risk event is used. 
With this approach, they enable risk calculation along continuous paths and better formalize robot interactions in the presence of uncertainty. Later work on lambda field mapping shows how it can be extended to three-dimensional spaces and integrated more closely with path planning \cite{randriamiarintsoa_risk-aware_2023}.

Beyond purely geometric maps, multi-channel representations accumulate evidence on additional attributes of the environment \cite{zhang_autonomous_2024, razavi_online_2025, canh_object-oriented_2024, gan_multitask_2023}. 
G-VOM \cite{overbye_g-vom_2021} computes hard and soft obstacle, slope, and roughness layers in a voxel grid for off-road driving, ForestTrav \cite{ruetz_foresttrav_2024} adds learned traversability estimates for overgrown forest terrain, while SPOMP \cite{miller_air-ground_2024} demonstrates the use of semantic maps as the shared representation for air--ground teams at field scale. 
Drawing from these prior works, SAFARI maintains a multi-channel map of evidence fields using a unified update step. 
We build on the lambda field formulation by adding a semantic hazard layer, uncertainty estimation, and dedicated layers for traversability, risk, and information gain.

\subsection{Mapping for Path Planning}

Point-cloud-based terrain analysis can extract useful information about the environment, such as slope, roughness, and step features that produce traversability costs to guide motion planning \cite{wermelinger_navigation_2016, fankhauser_probabilistic_2018, fan_step_2021}.
These geometric descriptors, however, are ambiguous in volumetric terrain, where, for example, traversable grass and a rigid obstacle of the same height are difficult to distinguish. 
Learned traversability estimators tackle this by predicting terrain properties beyond geometry from the same sensor data, such as rich voxel features in dense forests \cite{ruetz_foresttrav_2024} or self-supervised costs predicted from the robot's own experience in simulated environments \cite{frey_locomotion_2022}.

Probabilistic map representations expand the capabilities of geometric planners, allowing them to become risk-aware and more reactive in unknown environments \cite{lima_under-canopy_2024}. 
Rather than committing to a binary traversability decision, such planners reason about the uncertainty in the underlying estimate, for instance, by optimizing a conditional value-at-risk objective over probabilistic traversability distributions \cite{fan_step_2021}, or by enforcing safety constraints through control barrier functions defined over occupancy maps \cite{kim_safe_2025}. 
This is particularly valuable in unknown environments, where sparse or missing observations must be distinguished from confidently observed free space.

Semantic understanding, typically obtained from camera imagery, can further complement path planning solutions by providing secondary evidence for traversability assessment. Object-specific semantics allow planners to reason about geometrically ambiguous objects \cite{canh_object-oriented_2024}, and off-road semantic segmentation improves understanding of volumetric environments \cite{zhang_autonomous_2024, frey_fast_2023}. 
In search-and-rescue settings, semantic risk maps derived from visual and depth cues have been used to adapt waypoints online to the perceived danger of the surrounding terrain \cite{ulloa_bio-inspired_2026}.

These works establish that geometric and semantic information are complementary for robust planning in varied environments. 
SAFARI exposes both to the planner as continuous multi-channel evidence rather than as binary labels.

\subsection{Informative Path Planning}
Informative path planning selects robot trajectories to maximize the expected information gathered about the environment. Most commonly, map coverage is used as a proxy for information gain \cite{dai_fast_2020, wang_actively_2020, border_surface_2024, tankasala_beyond_2024, isler_information_2016, zhao_tdle_2023, zhou_fuel_2020, cao_autonomous_2021}. Some planners rely on frontier detection to select new viewpoints by the expected reduction of occupancy entropy \cite{dai_fast_2020}. Some greedily select the next-best-view (NBV) by trading information gain against the cost of reaching it \cite{wang_actively_2020, border_surface_2024}, an approach that can suffer from short-sighted behavior that longer horizon reasoning can mitigate \cite{tankasala_beyond_2024, isler_information_2016}. Others explicitly balance exploration and exploitation through hierarchical methods that define globally consistent goals for a lower-level navigation controller to reach \cite{zhao_tdle_2023, zhou_fuel_2020, cao_autonomous_2021}. One class of these methods dynamically divides the planning space into sub-regions and selects one to provide global guidance, and uses local indicators to target specific areas within each \cite{zhao_tdle_2023}. Others avoid explicit spatial partitioning: a global frontier structure is incrementally maintained and a sequence computed over frontier clusters before refining a local viewpoint sequence \cite{zhou_fuel_2020}, or a two-layer, multi-resolution representation of the environment is used to perform this same coarse-to-fine refinement \cite{cao_autonomous_2021}.

More recent work generalizes the gain definition beyond geometry to multiple information channels. Instead of aiming for the shortest mission time or maximum coverage, these methods consider the information gain, or uncertainty reduction. Papatheodorou et al. \cite{papatheodorou_finding_2023} jointly optimize free-space discovery, observation distance to background surfaces, and object reconstruction quality within a single objective. Semantic rewards can be introduced into roadmap-based exploration planners \cite{nguyen_semantic-aware_2026}, and viewpoint utilities can be formulated to unify geometric view diversity with open-vocabulary task relevance \cite{nagami_vista_2025}. Learned policies can internalize such multi-objective gains end-to-end for semantic inspection tasks \cite{malczyk_semantically-driven_2025}.

Our exploration objective follows a hierarchical structure: a global goal is selected at the frontier estimated to be most informative, and a local NBV planner executes discrete steps toward that goal while maximizing observed area along the way. Both levels operate directly on the channels of the lambda map.

\subsection{Autonomous Exploration Tasks \& Field Trials}
%
Autonomous exploration systems usually combine 3-D occupancy mapping, frontier selection, and informative viewpoint selection to guide navigation, but differ in whether they reason about semantics. 
One such system combines frontier estimation with sampling-based viewpoint selection for micro-aerial vehicles for occupancy-informed exploration using a stereo camera \cite{dai_fast_2020}. Another performs autonomous aerial spatial mapping of industrial areas by combining LiDAR-based SLAM with NBV planning \cite{border_osprey_2024}. Both demonstrate the benefits of combining mapping and navigation subsystems, but are limited to geometric mapping of feature-dense environments rather than large outdoor areas. Systems that do carry semantics divide into two groups: \textit{Object Finding} and \textit{Environment Assessment}.

\subsubsection*{Object Finding}
In object-goal navigation, a target object is specified before the mission begins, and locating it defines success. This formulation is well established indoors, where it is typically validated in simulation with limited real-robot testing \cite{qu_ippon_2024, zhao_decision-driven_2026, nagami_vista_2025}. Target localization also motivates search and rescue, where the target is a human victim. One system evaluated an autonomous UAV search for human targets in a real Mediterranean karst wilderness environment, using 78 human volunteers as search subjects \cite{dumencic_experimental_2025}. Search-and-rescue systems can also use continuing survey objectives, as discussed below. Our task requires surveying a designated region and reporting OPIs throughout exploration.

\subsubsection*{Environment Assessment}
In this category, we consider systems in which a more general target object class is provided, with a focus on cataloging and reporting the state of a scene. The DARPA Subterranean Challenge is the clearest example of this formulation, scoring systems on their ability to explore unknown subterranean areas while reporting artifacts within them. The CERBERUS team combined legged and flying robots with unified exploration planning and learning-based detection under the supervision of a single operator \cite{tranzatto_team_2022}. Robinson et al.~\cite{robinson_human-robot_2022} compare fully autonomous search against human-robot teaming across 16 real-world missions adapted from the challenge. The missions use fixed time budgets and continue after an artifact is found. Beyond the subterranean domain, assessment missions have been deployed across a range of application areas, including forest surveying \cite{mattamala_building_2025}, industrial structure reconstruction \cite{wang_actively_2020}, and multi-agent geometric and semantic mapping \cite{liu_active_2023}. Notably, litter-cataloging missions have surveyed and reported waste from patrol imagery \cite{kraft_autonomous_2021}, and planned efficient collection routes once waste was located \cite{amatucci_vero_2024}.

A closer task formulation comes from radiological inspection, where the measured quantity is both the mission deliverable and a possible threat to the platform. Radiation measurements are converted to a costmap layer, which is used as evidence to avoid certain areas, so that the robot does not accumulate absorbed dose \cite{groves_robotic_2021, west_real-time_2022}. Later systems extend this to alpha- and beta-contamination and fuse the radiation layer with a terrain layer for autonomous surveying \cite{nouri_rahmat_abadi_carma_2023}. The measurements constrain where the robot may go, but never inform the robot where it should look next.

Reporting of explosive remnants of war is the assessment task closest to ours in stakes, but robotic mine detection has developed along a largely separate track, dominated by sensor-specialized detection: proximal metal detectors or ground-penetrating radar on ground platforms \cite{portugal_deploying_2014, dasgupta_comrade_2015}, or thermal and magnetometric sensing from the air \cite{bajic_uav_2023}. There, the platform sweeps an area, and the collected data is interpreted offline, so no detection outcome influences where the platform goes next.

Large-scale field validation of autonomous survey systems is well established. 
CERBERUS reports extended deployments across tunnel, urban, and cave environments in preparation for the SubT Challenge \cite{tranzatto_team_2022}.
Forest inventory \cite{mattamala_building_2025} and volcanic gas monitoring \cite{richter_large-scale_2026} demonstrate legged autonomy over large areas in unstructured terrain, and search-and-rescue studies are often evaluated across multiple real missions \cite{robinson_human-robot_2022, dumencic_experimental_2025}. 
What these campaigns do not report is how a hazard channel behaves once it is coupled to motion, since in each case, detection and traversal remain separate. Our deployment, therefore, targets that coupling directly: we characterize it across three outdoor sites and 11 missions, varying the parameters that arbitrate between observing an OPI more closely and avoiding it.

Across these systems, detections play different roles. In object-goal navigation, locating the specified target defines success \cite{qu_ippon_2024, zhao_decision-driven_2026, nagami_vista_2025}. In assessment formulations, objects are reported during continued exploration \cite{robinson_human-robot_2022, tranzatto_team_2022, kraft_autonomous_2021, liu_active_2023}. In radiological inspection, the measurement repels the platform without indicating where to look next \cite{groves_robotic_2021, west_real-time_2022, nouri_rahmat_abadi_carma_2023}.
In the remaining systems, detections are scored after the mission \cite{bajic_uav_2023} or kept outside the navigation stack \cite{richter_large-scale_2026}. Our task differs in that an OPI must attract and repel at once. We therefore accumulate detections into the same risk field that the planner reads, and both planning levels compute their objectives from a single representation.

These differences complicate a direct system-level comparison.
Adapting the surveyed systems to jointly use semantic hazard evidence for further inspection and risk-aware motion planning would require changes to the mapping and planning components under test. Additionally, many of the previously discussed systems are not publicly available. We therefore evaluate our contributions at the component level through field ablations using the same hardware at the same sites, and assess the integrated system through field deployment.

\definecolor{methodadditioncolor}{RGB}{0,105,75}
\definecolor{methodremovalcolor}{RGB}{190,0,0}
\definecolor{methodnotecolor}{RGB}{0,0,128}
\providecommand{\methodadd}[1]{\textcolor{methodadditioncolor}{#1}}
\providecommand{\methodremove}[1]{\textcolor{methodremovalcolor}{#1}}
\newenvironment{methodaddition}{\begingroup\color{methodadditioncolor}}{\endgroup}
\newenvironment{methodremoval}{\begingroup\color{methodremovalcolor}}{\endgroup}
\providecommand{\methodnote}[1]{%
    \par\noindent\textcolor{methodnotecolor}{\textbf{Author note:} #1}\par
}

\begin{figure*}[t]
    \centering
    \includegraphics[width=1.0\linewidth]{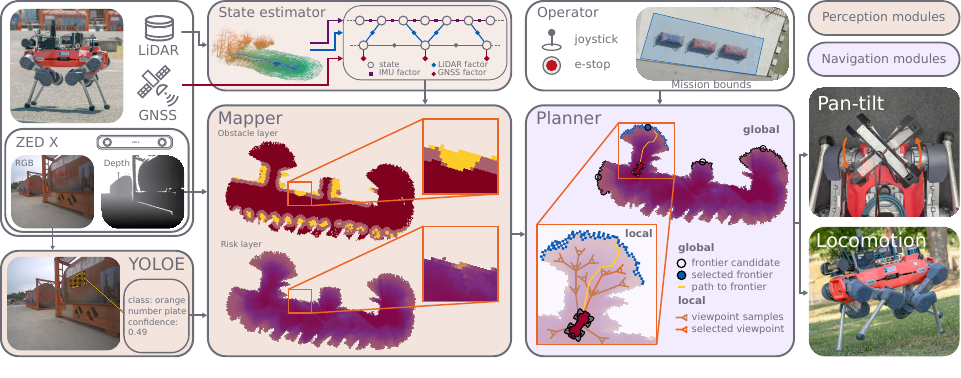}
    \caption{System overview diagram of the SAFARI framework. It depicts the flow of information through the detection, mapping and navigation modules, as well as the operating concept for each. It combines real data (sensor data, detector output, map layers) and illustrations (robot, viewpoints, paths) for a more comprehensive depiction.}
    \label{fig:system_overview}
\end{figure*}

\section{PRELIMINARIES}
\label{sec:preliminaries_draft}

\subsection{Problem Formulation}
\label{sec:problem_formulation}

We consider a mobile robot tasked with autonomously surveying an area $\Omega \subset \mathbb{R}^{2}$, whose boundary is provided before deployment. 
No prior geometric or semantic map of the interior of $\Omega$ is assumed to be available. The environment may contain untraversable terrain, geometric obstacles, occlusions, and OPIs belonging to a set of hazard classes $\mathcal{H}$.

The robot configuration used for planning at time $t$ is denoted by $\mathbf{x}_t \in SE(2)$, where motion is represented in the horizontal plane, while the camera pose used for perception is represented in $SE(3)$. Its onboard camera pose in the world frame is denoted by $\mathbf{T}^{W}_{C,t} \in SE(3)$. During operation, the robot receives an RGB image $I_t$, a registered depth measurement $D_t$, and a localization estimate $\hat{\mathbf{T}}^{W}_{C,t}$. We collect these inputs in the observation
\begin{equation}
    \mathbf{z}_t =
    \left(
        I_t,
        D_t,
        \hat{\mathbf{T}}^{W}_{C,t}
    \right).
\end{equation}

Using the sequence of observations $\mathbf{z}_{0:t} = \{\mathbf{z}_0,\ldots,\mathbf{z}_t\}$, the robot must construct and maintain a spatial map $\mathcal{M}_t$ representing both the physical structure of the environment and the accumulated evidence regarding the presence of hazards.

A positive hazard observation consists of a semantic detection that can be associated with valid depth, transformed into the world frame using $\hat{\mathbf{T}}^{W}_{C,t}$, and associated with a location in $\Omega$. A negative hazard observation is assigned only to a visible map location at which a particular OPI-size hypothesis is predicted to be detectable under the observation model, but no corresponding detection was made. The absence of a detection in $I_t$ does not by itself constitute negative evidence.

At each planning iteration, the system generates a feasible robot trajectory
\begin{equation}
    \boldsymbol{\xi}
    =
    \left(
        \mathbf{x}_{t},
        \mathbf{x}_{t+1},
        \ldots,
        \mathbf{x}_{t+N}
    \right).
\end{equation}
that increases the amount of hazard- and geometry-relevant information collected from $\Omega$. 
A feasible trajectory must remain inside the mission boundary and satisfy the currently estimated geometric and traversability constraints, while proximity to hazard sources is considered during trajectory evaluation. 
The mission-level objective is to reduce the area that is not yet observed sufficiently while limiting travel cost and exposure to unsafe regions. 
The required output is a persistent hazard map that identifies regions containing hazard evidence, regions supported by sufficient negative evidence, and regions for which the available observations remain inconclusive.

\subsection{Lambda Field Representation}
\label{sec:lambda_field_prelim_draft}

This section summarizes the principles of the lambda field introduced by Laconte et al.~\cite{laconte_novel_2021}. The main difference between a classical occupancy grid and a lambda field is the quantity assigned to each cell. 

A conventional occupancy grid stores a per-cell probability that the cell is occupied, which is tied to the chosen resolution. A lambda field instead stores a density, namely the expected number of events such as collisions per unit area, from which the probability of an event in any region follows by integration.

Let a planar domain be discretized into cells $c_i \in \mathcal{C}$ of area $\Delta a$. Each cell is assigned an intensity
\begin{equation}
    \lambda_i \in \mathbb{R}_{\geq 0}.
\end{equation}
A value $\lambda_i=0$ indicates that no event is expected in the cell, whereas a large value indicates a high event intensity per unit area. For a region $A\subseteq\mathbb{R}^2$, the probability that at least one event will occur in that region is
\begin{equation}
    P(\mathrm{event}\mid A)
    =
    1-\exp\left(-\int_A \lambda(\mathbf{p})\,d\mathbf{p}\right).
    \label{eq:lambda_area_probability_prelim_draft}
\end{equation}
Assuming a piecewise-constant field, the probability of a collision along a path $P$ that intersects the set of cells $\mathcal{C}_P$ is approximated by
\begin{equation}
    P(\mathrm{coll}\mid P)
    =
    1-
    \exp\left(-\Delta a
    \sum_{c_i\in\mathcal{C}_P}\lambda_i\right).
    \label{eq:lambda_path_collision_prelim_draft}
\end{equation}
The formulation, therefore, represents collision-event intensity per unit area rather than assigning a collision probability directly to each cell. This distinction enables the collision risk to be calculated by integrating the area swept by the robot's body over the desired path.

In the formulation proposed by Laconte et al.~\cite{laconte_novel_2021}, a two-dimensional LiDAR ray provides two types of evidence. Cells traversed before the endpoint are counted as misses because the beam passed through them without a collision. Cells within an error region around the endpoint are counted as hits because the surface that interacted with the beam is assumed to lie within this neighborhood. Let $m_i$ and $h_i$ denote the accumulated miss and hit evidence for cell $c_i$, respectively. Let $\mathcal{E}$ denote the measurement error region, within which the true surface is assumed to lie, and let 
$\epsilon = |\mathcal{E}|$ denote its area. 
For a given $\epsilon$, the maximum-likelihood intensity estimate is
\begin{equation}
    \lambda_i
    =
    \frac{1}{\epsilon}
    \ln\left(1+\frac{h_i}{m_i}\right).
    \label{eq:lambda_estimation_prelim_draft}
\end{equation}

Consequently, cells repeatedly crossed without a ray endpoint receive a low intensity, while cells that repeatedly contain ray endpoints receive a high intensity. The factor $1/\epsilon$ converts the hit-to-miss evidence ratio into an intensity per unit area. A larger error region distributes positive evidence over a larger area and therefore produces a lower local intensity. This parameter depends on the capabilities of the sensor used.

\section{METHOD}
\label{sec:method_draft}
\subsection{Overview}
An overview of the proposed system is shown in \Cref{fig:system_overview}. The framework is organized as a closed-loop pipeline consisting of three principal modules: semantic RGB-D perception, probabilistic multi-layer mapping, and coverage planning.

The perception stage runs an open-vocabulary segmentation model on each RGB image and associates the resulting masks with registered depth, producing spatially anchored detections. 
The mapping stage accumulates this output in a lambda field \cite{laconte_novel_2021}, where depth measurements provide geometric hit and miss evidence, detections with valid depth provide positive hazard evidence, and visible regions in which an OPI of a given size would have been detectable but was not provide negative hazard evidence. 
Rather than committing to binary labels (occupied or free, hazardous or safe), the mapper retains this evidence and derives from it the traversability, hazard-intensity, and observation-uncertainty layers described in \Cref{sec:mapping_draft}.
This preserves the distinction between an area that appears hazard-free and one that has not yet been sufficiently observed.
The planning stage queries these layers to select informative, reachable, and low-risk trajectories. 
The executed motions produce new observations that update the map and, in turn, the utility of future trajectories, allowing the robot to adapt to newly discovered obstacles, terrain, and hazards without a prior map.

\begin{figure*}
    \centering
    \includegraphics[width=1.0\linewidth]{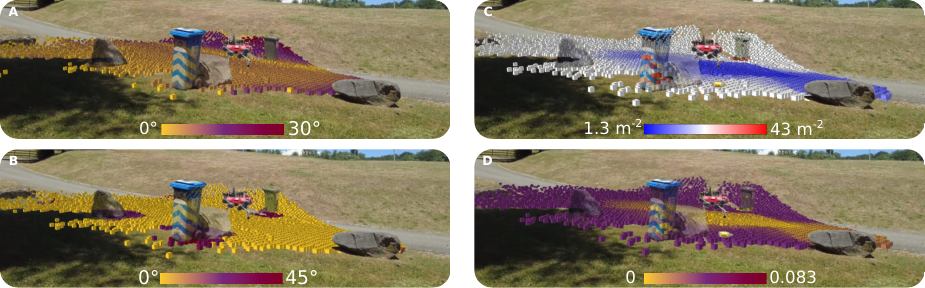}
    \caption{Map and traversability layers overlaid on the mission footage. Panels (A) and (B) show the geometric layer, while panels (C) and (D) show the hazard layer. \textbf{(A)} The terrain-grade estimate $\theta_i^{\mathrm{grade}}$, representing the large-scale inclination of the estimated support surface. \textbf{(B)} The residual-slope estimate $\theta_i^{\mathrm{res}}$, highlighting local height discontinuities and protrusions after removing the broader terrain grade. \textbf{(C)} The hazard lambda field intensity $\lambda^{\mathrm{haz}}_{i,k}$, representing the spatial intensity derived from the accumulated positive and negative risk evidence for the active size hypothesis. \textbf{(D)} The corresponding Beta variance $V_{i,k}$, used as an uncertainty measure for the accumulated hazard-evidence rate.}
    \label{fig:mission_overlay}
\end{figure*}

\subsection{Semantic RGB-D Observation Generation}
\label{sec:observation_generation_draft}

The semantic perception module takes the raw RGB-D measurements and converts them to spatial risk observations to be used by the mapping framework. The RGB image is processed by an open-vocabulary YOLOE segmentation model \cite{wang_yoloe_2025} with text prompts, exported to TensorRT. Given an image, the model returns a confidence score, an image-space region of interest, and a binary segmentation mask for each accepted detection. Segmentation masks are used by the mapper to sample depth measurements from pixels likely to belong to the detected object instead of associating the entire bounding box with the hazard class. Detections below a confidence threshold are discarded, and the remaining masks are transformed back to the original image geometry.

The accepted detection pixels are published with their confidence values and synchronized with an organized depth point-cloud. Each sampled detection pixel is reconstructed from the corresponding depth value, transformed into the world frame using $\hat{\mathbf{T}}^W_{C,t}$, and assigned to a map cell. Outlying or invalid depth samples are discarded. If an image-space detection contains no valid depth samples, it is not allowed to create positive hazard evidence. This prevents a purely image-space false detection from becoming a persistent spatial hazard. This separation is necessary because individual visual detections are uncertain. Singular frames may contain false positives, missed detections, motion blur, depth errors, or challenging illumination. Treating the detector as an evidence source allows evidence collected from multiple observations and viewpoints to be accumulated by the mapper accounting for their quality and visibility conditions instead of applying the outcome of a single frame as a hard label.

\subsection{Probabilistic Multi-Layer Mapping}
\label{sec:mapping_draft}

The mapping framework maintains the persistent spatial representation of the mission area by accumulating geometric and semantic risk evidence through the mission. It represents geometric occupancy, traversability, as well as the semantic evidence concerning OPIs. Information about traversable space, potential danger zones, and regions where additional measurements are expected to improve the quality or extent of the map is derived from the map to be used for path planning and exploration.

\subsubsection{Use of Lambda Fields}
\label{sec:lambda_field_use_draft}

Our proposed mapper uses the lambda field formulation \cite{laconte_novel_2021}, which represents the spatial intensity of events estimated from the accumulated hit and miss evidence and allows the consistent calculation of probabilities for arbitrary map resolutions.

This formulation uses geometric and semantic risk information and is maintained in two layers. The geometric layer represents collision-related intensity associated with physical structures, while the hazard layer represents the spatial intensity of accumulated evidence associated with hazardous objects. Both layers share the same mathematical basis but differ in their observation models and their interpretation. \Cref{fig:mission_overlay}-A and -B shows the components of the geometric layer, while \Cref{fig:mission_overlay}-C and -D shows the hazard layer during a mission.

For each persistent cell $c_i$, the mapper stores the state
\begin{equation}
\mathbf{s}_i =
\left(
H_i^{\mathrm{geom}},
M_i^{\mathrm{geom}},
H_i^{\mathrm{haz}},
M_{i,1}^{\mathrm{haz}},
\ldots,
M_{i,K}^{\mathrm{haz}},
\hat{z}_i
\right),
\label{eq:cell_state_draft}
\end{equation}

where $H_i^{\mathrm{geom}}$ and $M_i^{\mathrm{geom}}$ are the geometric hit and miss evidence, $H_i^{\mathrm{haz}}$ is the positive hazard evidence accumulated from detections, and $M_{i,k}^{\mathrm{haz}}$ is the negative hazard evidence accumulated under object-radius hypothesis $R_k$ for $k=1,\ldots,K$. The positive statistic is shared across all size hypotheses, because a detection is evidence of a hazard irrespective of the hypothesis under which the cell is being cleared, whereas a non-detection only clears the cell of objects large enough to have been seen. Finally, $\hat{z}_i$ is the estimated ground-support height of the cell. Where the layer is unambiguous from context, the superscript is omitted and we write $H_i$ and $M_{i,k}$ for the hazard statistics.

The stored state evidence is then used to derive the geometric and semantic lambda field intensities, along with additional layers used by the planner. These include risk-evidence uncertainty, expected variance reduction, and traversability.

\subsubsection{Geometric Evidence Accumulation}
\label{sec:geometric_mapping_draft}

Existing lambda field implementations accumulate evidence from sparse LiDAR scans either on a 2-D \cite{laconte_novel_2021} or 2.5-D grid \cite{randriamiarintsoa_risk-aware_2023}. 
The RGB-D sensor used here instead provides dense depth measurements that cannot be reduced to a single planar scan without discarding most of the available spatial structure. Directly projecting every point onto a horizontal grid would collapse ground points, obstacles, and overhanging structures into the same cell. 
Similarly, denoting free space only in the horizontal plane would not distinguish whether a visibility ray passed above the ground, through the robot-relevant obstacle band, or through irrelevant space. 
This could make the map overly conservative in some regions and overconfident in others.
The geometric mapping layer, therefore, discretizes the third dimension locally. Around the current sensor pose, the mapper constructs a local voxel representation aligned with the persistent two-dimensional lambda grid. Each valid RGB-D point-cloud sample is treated as an endpoint measurement in this local volume. A three-dimensional ray is traced from the camera origin to the depth sample; voxels traversed by the ray receive geometric miss evidence, while a small error volume surrounding the endpoint receives geometric hit evidence.

\begin{table}[t]
    \centering
    \resizebox{\linewidth}{!}{%
    \begin{tabular}{l l l}
        \toprule
        Symbol & Description & Value \\
        \midrule
        \multicolumn{3}{l}{\textit{Geometric layer}}\\
        $\epsilon_{\mathrm{geom}}$ & Error-region footprint area & $0.10\times0.10\,\mathrm{m}^2$ \\
        $\alpha_0^{\mathrm{geom}}$ & Hit prior count & $0.5$ \\
        $\beta_0^{\mathrm{geom}}$ & Miss prior count & $25.0$ \\
        \midrule
        \multicolumn{3}{l}{\textit{Hazard layer}}\\
        $A_{\mathrm{haz}}$ & Hazard error-region area & $0.25\times0.25\,\mathrm{m}^2$ \\
        $\alpha_0$ & Hit prior count & $1.0$ \\
        $\beta_0$ & Miss prior count & $1.0$ \\
        $K$ & Number of size hypotheses & $5$ \\
        $R_k$ & OPI-radius hypothesis & $\{0.05,0.10,0.20,0.60,1.20\}\,\mathrm{m}$ \\
        $p_{\min}$ & Minimum apparent diameter & $18\,\mathrm{px}$ \\
        $p_{\mathrm{margin}}$ & Apparent-diameter margin & $22\,\mathrm{px}$ \\
        $f_x$ & Focal length & from intrinsics \\
        $w_0$ & Base miss weight & $0.9$ \\
        $\eta_k$ & Per-hypothesis miss multiplier & $\{0.5,0.8,1.0,1.0,1.0\}$ \\
        $v_{\mathrm{ref}}$ & Motion-attenuation reference speed & $0.75\,\mathrm{m/s}$ \\
        $w^{\min}_{\mathrm{motion}}$ & Motion-attenuation floor & $0.10$ \\
        $w_{\mathrm{cap}}$ & Same-frame miss sum cap & $1.0$ \\
        $E_{\max}$ & Evidence clamp radius & $80$ \\
        $\lambda_{\max}$ & Intensity saturation bound & $5000\,\mathrm{m}^{-2}$ \\
        $U_{\min}$ & Uncertainty score threshold & $0.5$ \\
        -- & Miss-ray end margin & $0.08\,\mathrm{m}$ \\
        -- & Visibility sampling range & $0.75$--$6.0\,\mathrm{m}$ \\
        \bottomrule
    \end{tabular}%
    }
    \footnotesize
    \setlength{\tabcolsep}{4pt}
    \caption{Mapping parameters used.}
    \label{tab:mapping_params}
\end{table}

This procedure produces volumetric hit and miss evidence, but the planner does not consume a dense three-dimensional grid. For the geometric layer, voxel evidence over a ground-relative height interval is summed column-wise into the persistent statistics $H_i^{\mathrm{geom}}$ and $M_i^{\mathrm{geom}}$ of the corresponding two-dimensional cell, and the geometric intensity is recovered with the same estimator used for the hazard layer,
\begin{equation}
    \lambda^{\mathrm{geom}}_i
    =\min\left(
        \frac{1}{\epsilon_{\mathrm{geom}}}
        \log\left(1+\frac{H_i^{\mathrm{geom}}}{M_i^{\mathrm{geom}}}\right),
        \ \lambda_{\max}
    \right),
    \label{eq:geometric_lambda_estimation_draft}
\end{equation}
where $\epsilon_{\mathrm{geom}}$ is the planar area of the geometric error region and $\lambda_{\max}$ is a saturation bound. Since the column sum reduces evidence collected in three dimensions onto a planar cell, $\epsilon_{\mathrm{geom}}$ is specified as an area so that $\lambda^{\mathrm{geom}}_i$ keeps units of inverse area and stays comparable with Eq.~\eqref{eq:lambda_path_collision_prelim_draft}. Because the statistics are initialized at their priors, a cell that has never been measured does not report zero intensity but the prior value $\smash{\epsilon_{\mathrm{geom}}^{-1}\log(1+\alpha_0^{\mathrm{geom}}/\beta_0^{\mathrm{geom}})}$; the priors in \Cref{tab:mapping_params} are chosen so that this floor is small, and unobserved space is separated from observed free space by the free-space mask rather than by the intensity itself. The mapper additionally supports injecting a virtual obstacle intensity around tracker-confirmed hazards, which would couple the OPI layer back into traversability.

This discrete volumetric structure is preserved only locally. The navigation and traversability states are projected into the persistent 2-D grid for ease of use and long-term storage. For each column, the lowest cluster of reliable surface observations is used to update a ground-height estimate and its confidence. Voxel evidence is then interpreted relative to this ground-support estimate. Using this estimate as a starting point, the detections within the same vertical range of the robot's body are classified as either impassable ground protrusions or obstacles, such as large rocks, trees, or humans. The ground-height estimate is also used to assess the local and broader slope of the terrain. From these computations, a traversability map of the environment is extracted and stored in the persistent flat representation to be used by the path planner.

The geometric and hazard layers therefore share an evidence-accumulation principle, but differ in what their evidence represents, how it is filtered, and how the resulting information is compressed into two-dimensional map layers.

\begin{figure*}[t]
    \centering
    \includegraphics[width=0.98\linewidth]{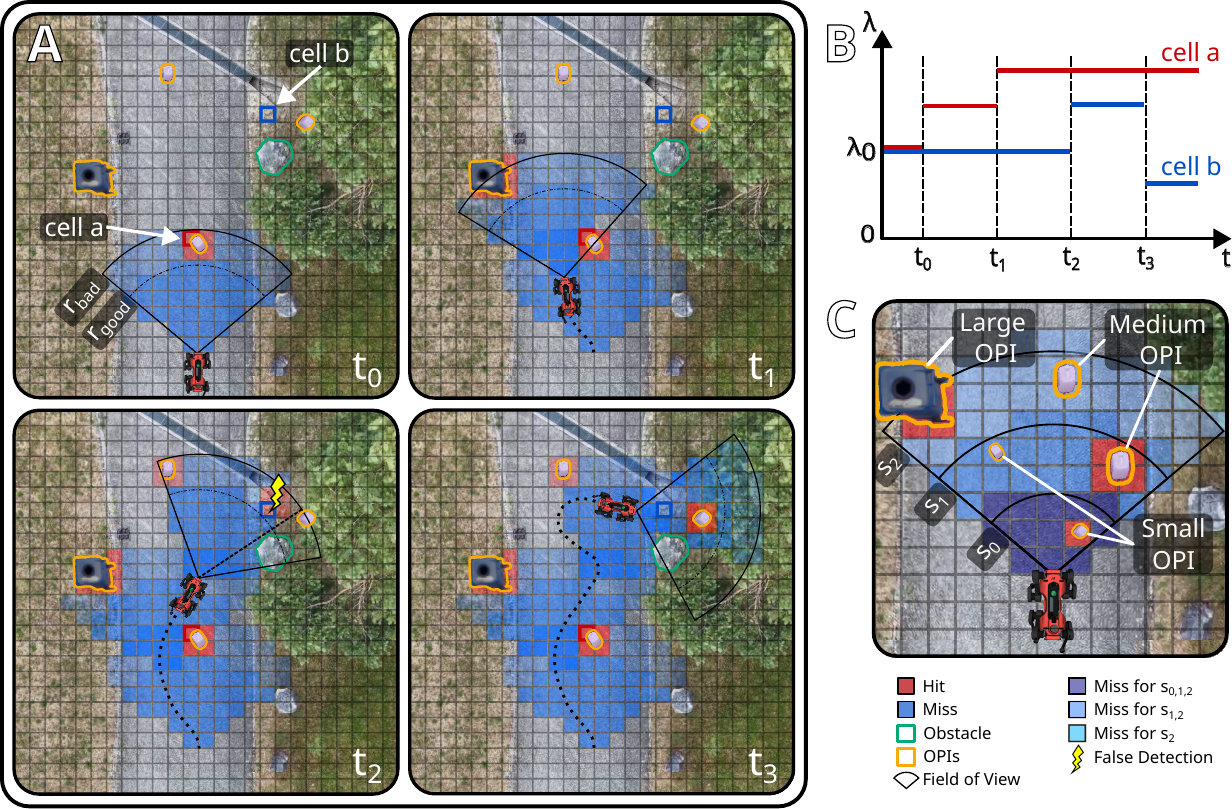}
    \caption{Figure demonstrating the OPI evidence accumulation of the lambda map. \textbf{(A)} Updates to the OPI layer of the lambda map at different time steps of a mission. At $t_2$, a false detection is simulated to show how the mapper recovers from it. \textbf{(B)} Changes in the $\lambda_i$ values of two selected map cells. \textbf{(C)} The expected OPI size mechanism. Small and medium OPIs are not detected beyond the distances $s_0$ and $s_1$, respectively.}
    \label{fig:mapper_acc}
\end{figure*}

\subsubsection{OPI Evidence Accumulation}
\label{sec:hazard_mapping_draft}
The stored statistics are initialized with positive pseudo-counts $\alpha_0$ and $\beta_0$. Their state after accumulating observations up to time $t$ can be written as
\begin{equation}
    H_i = \alpha_0 + \sum_{\tau=1}^{t} w^+_{i,\tau},
    \qquad
    M_{i,k} = \beta_0 + \sum_{\tau=1}^{t} w^-_{i,k,\tau}.
    \label{eq:hazard_evidence_draft}
\end{equation}
The increments are non-negative, potentially fractional evidence weights. A detection updates the shared hit statistic $H_i$, whereas visibility without a detection updates the miss statistic $M_{i,k}$ associated with OPI-radius hypothesis $R_k$.

The OPI-size hypothesis is illustrated in \Cref{fig:mapper_acc}-C. A larger hypothesis extends the modeled detectability ranges and increases the weight of negative evidence at a given distance. This can assign negative evidence where smaller OPIs are not reliably detected. A smaller hypothesis reduces the full-weight and partial-weight ranges defined below, which can require closer viewpoints or more observations to meet the observation threshold.

Positive hazard evidence is generated only from the valid, depth-supported segmentation samples described in \Cref{sec:observation_generation_draft}. Each reconstructed sample is transformed into the map frame and accumulated as hazard hit evidence $H_i$ in the corresponding cell, weighted by the confidence of the sample it came from. Writing $\mathcal{S}_{i,t}$ for the set of depth-backed mask samples of frame $t$ that fall in cell $c_i$, and $\gamma_{s}$ for the per-sample detection confidence,
\begin{equation}
    w^+_{i,t}=\sum_{s\in\mathcal{S}_{i,t}}\max\left(0,\gamma_{s}\right).
    \label{eq:hazard_hit_weight_draft}
\end{equation}
Cells that receive positive evidence in a frame are excluded from that frame's miss update, so a detected surface is never simultaneously cleared.

Negative OPI evidence is generated from simulated visibility rays obtained using the depth information. The mapper extracts sparse visibility samples from the synchronized depth cloud and traces rays from the camera origin toward the observed surface points. For every hazard-size hypothesis $k$, cells traversed by a ray receive miss evidence $M_{i,k}$ only where an object of the corresponding radius should have been visually detectable given the camera intrinsics and apparent-size thresholds. 

The ray terminates before the measured surface by the hypothesized object radius and a small margin so that a hazard directly in front of an observed surface is not incorrectly cleared.

Detectability is estimated from the apparent image diameter of a hazard. For camera focal length $f_x$, the distance at which an OPI of radius $R_k$ reaches the preferred and minimum apparent diameters is
\begin{equation}
    r^{\mathrm{good}}_k
    = \frac{2f_xR_k}{p_{\min}+p_{\mathrm{margin}}},
    \qquad
    r^{\mathrm{bad}}_k
    = \frac{2f_xR_k}{p_{\min}}.
    \label{eq:hazard_detectability_ranges_draft}
\end{equation}
The geometric detectability factor is one up to $r^{\mathrm{good}}_k$, decreases linearly between $r^{\mathrm{good}}_k$ and $r^{\mathrm{bad}}_k$, and is zero at and beyond $r^{\mathrm{bad}}_k$,
\begin{equation}
    p_{\mathrm{det}}(r,k)=
    \begin{cases}
        1, & r\leq r^{\mathrm{good}}_k,\\[2pt]
        \dfrac{r^{\mathrm{bad}}_k-r}{r^{\mathrm{bad}}_k-r^{\mathrm{good}}_k},
            & r^{\mathrm{good}}_k<r<r^{\mathrm{bad}}_k,\\[6pt]
        0, & r\geq r^{\mathrm{bad}}_k .
    \end{cases}
    \label{eq:hazard_detectability_factor_draft}
\end{equation}
After applying the depth-validity and ray-visibility checks, an accepted negative update is weighted by $w^-$, which is defined as
\begin{equation}
    w^-_{i,k,t}
    = w_0\,\eta_k\,p_{\mathrm{det}}(r_{i,t},k)\,
      w_{\mathrm{motion},t},
    \label{eq:hazard_miss_weight_draft}
\end{equation}
where $w_0$ is the base miss weight, $\eta_k\in(0,1]$ is a per-hypothesis multiplier that discounts clearing claims made under the smallest radii, $r_{i,t}$ is the range from the camera to cell $c_i$, and $w_{\mathrm{motion},t}$ down-weights evidence acquired during rapid motion that might introduce motion-blur. The latter decays with base speed relative to a reference speed $v_{\mathrm{ref}}$, carries an additional penalty at long range, and is floored at $w_{\mathrm{motion}}^{\min}$. To prevent a dense depth cloud from being interpreted as many independent non-detections, repeated same-frame miss contributions to a cell are combined with a maximum rather than accumulated; a capped sum with limit $w_{\mathrm{cap}}$ is available as an alternative. Positive evidence is not combined in this way: all depth-backed detection samples falling in a cell within a frame are summed, so a single large mask can deposit more evidence in one frame than a single visibility ray can remove.

Panel (A) of \Cref{fig:mapper_acc} illustrates the ranges $r^{\mathrm{good}}_k$ and $r^{\mathrm{bad}}_k$. Negative updates are additionally limited by visibility and the configured ray range.

The stored evidence is bounded jointly. 

Whenever
\begin{equation}
    {\max_k\sqrt{H_i^2+M_{i,k}^2}} > E_{\max},
\end{equation}
where $E_{\max}$ is a configured limit,
the shared hit statistic and all size-dependent miss statistics are scaled by the single common factor
\begin{equation}
    s_i = \frac{E_{\max}}{\max_k\sqrt{H_i^2+M_{i,k}^2}},
    \label{eq:evidence_clamp_draft}
\end{equation}
which leaves every ratio $H_i/M_{i,k}$, and therefore every intensity in Eq.~\eqref{eq:hazard_lambda_estimation_draft} and every rate in Eq.~\eqref{eq:hazard_evidence_rate_draft}, exactly invariant. The clamp also bounds the total evidence $H_i+M_{i,k}$, moderating the reduction of $V_{i,k}$ and $\Delta V_{i,k}$ at a fixed evidence rate $q_{i,k}$. The residual uncertainty and observation demand depend on the balance of evidence and can still approach zero as $q_{i,k}$ approaches zero or one. We use this deliberately to limit confidence gains from correlated measurements taken from similar viewpoints, so $\Delta V_{i,k}$ should be read as an observation-demand signal under bounded evidence rather than as the variance reduction of an unbounded estimator.

The resulting $\lambda$ intensity for cell $c_i$ and size hypothesis $k$ is
\begin{equation}
    \lambda^{\mathrm{haz}}_{i,k}
    =
    \frac{1}{A_{\mathrm{haz}}}
    \log\left(1+\frac{H_i}{M_{i,k}}\right),
    \label{eq:hazard_lambda_estimation_draft}
\end{equation}
where $A_{\mathrm{haz}}$ is the area of the hazard error-region. A detection increases the hazard intensity locally, whereas repeated depth-supported observations without a detection reduce the intensity in cells where an object of size $k$ should have been observable. As for the geometric layer, the statistics are initialized at their priors, so a cell that has never been measured reports the prior intensity $\smash{A_{\mathrm{haz}}^{-1}\log(1+\alpha_0/\beta_0)}$ rather than zero. With the uniform prior $\alpha_0=\beta_0$ used here this corresponds to $q_{i,k}=1/2$, that is, an unmeasured cell is treated as maximally uncertain rather than as hazard-free. Intensity alone therefore cannot separate an unobserved cell from a measured one with a balanced evidence ratio, which is the reason the uncertainty measure of \Cref{sec:hazard_uncertainty_draft} is maintained alongside it.

The mission specifies a single operating radius $R^{*}$, and the miss statistic at that radius is obtained by linear interpolation between the two bracketing knots $R_k\leq R^{*}\leq R_{k+1}$,
\begin{equation}
    M_{i,k^{*}}=(1-t)\,M_{i,k}+t\,M_{i,k+1},
    \qquad
    t=\frac{R^{*}-R_k}{R_{k+1}-R_k},
    \label{eq:hazard_knot_interpolation_draft}
\end{equation}
and clamped to the end knots outside that range. All quantities carrying the index $k^{*}$ in the following are evaluated with this interpolated statistic and the shared $H_i$.

It is useful to isolate the evidence ratio
\begin{equation}
    q_{i,k}=\frac{H_i}{H_i+M_{i,k}},
    \qquad
    \lambda^{\mathrm{haz}}_{i,k}
    =-\frac{1}{A_{\mathrm{haz}}}\log(1-q_{i,k}).
    \label{eq:hazard_evidence_rate_draft}
\end{equation}
Here, $q_{i,k}$ is the mapper's positive-evidence rate under size-hypothesis $k$ and not the probability that a hazard physically exists in the cell. It estimates the probability of observing a hit in the next measurement. $\lambda^{\mathrm{haz}}_{i,k}$ is an intensity per unit area. A probability is obtained only after integrating this intensity over a specified area, as described in Eq.~\eqref{eq:lambda_area_probability_prelim_draft}. \Cref{fig:mission_overlay}-C shows the $\lambda^{\mathrm{haz}}_{i,k}$ during a mission.

\subsubsection{Hazard Belief and Uncertainty}
\label{sec:hazard_uncertainty_draft}

The hazard layer additionally provides an uncertainty measure used for information-gain calculations. The mapper interprets the accumulated hazard hit and miss evidence for cell $c_i$ and size hypothesis $k$ as the parameters of a Beta distribution. Specifically, the mapper models
\begin{equation}
    Q_{i,k}\sim\operatorname{Beta}(H_i,M_{i,k}),
    \label{eq:beta_evidence_model_draft}
\end{equation}
because $H_i$ and $M_{i,k}$ already include the configured pseudo-counts in
Eq.~\eqref{eq:hazard_evidence_draft}. Its mean is $q_{i,k}$ and its variance is
\begin{equation}
    V_{i,k}
    =\operatorname{Var}(Q_{i,k})
    =\frac{H_iM_{i,k}}
    {(H_i+M_{i,k})^2(H_i+M_{i,k}+1)}.
    \label{eq:beta_variance_revised_draft}
\end{equation}

The expected marginal reduction in this variance is used as the hazard information measure. With $q_{i,k}=H_i/(H_i+M_{i,k})$, the expected reduction is
\begin{equation}
\begin{aligned}
    \Delta V_{i,k}={}&V(H_i,M_{i,k}) \\
    &-q_{i,k}V(H_i+1,M_{i,k}) \\
    &-(1-q_{i,k})V(H_i,M_{i,k}+1),
\end{aligned}
\label{eq:expected_variance_reduction_revised_draft}
\end{equation}
where $V(\cdot,\cdot)$ denotes the variance in Eq.~\eqref{eq:beta_variance_revised_draft}. Actual mapper updates may have weights below one, so $\Delta V_{i,k}$ is used as a consistent planning estimate rather than a literal prediction of the next frame's variance. \Cref{fig:mission_overlay}-D shows the distribution of this measure during a mission.

This quantity identifies cells for which an additional valid hazard observation is expected to change the current belief. It is used as an observation-demand measure for planning rather than as a binary statement that a region is safe or hazardous. 

\subsubsection{Traversability and Map Layers for Planning}
\label{sec:planner_maps_draft}

The exploration planner requires a representation of the regions in which the robot can navigate safely. The mapper computes this information using a ground estimate and the geometric and hazard layers. 
For this, each cell maintains an estimate of ground height, the confidence associated with this estimate, and a terrain label. Depth measurements continuously update these quantities. Regions whose height is consistent with their neighbors are classified as ground, while sufficiently large discontinuities are classified as obstacles. Ground slope is estimated from the height estimates of neighboring cells \cite{randriamiarintsoa_risk-aware_2023}. Cells whose slope exceeds a threshold configured according to the capability of the robot are marked as untraversable.

To distinguish a sustained terrain inclination from a local height discontinuity, the selected traversability configuration decomposes the support surface into a broad terrain grade and a short-baseline residual slope. For every cell $c_i$ with a reliable support-height estimate $\hat{z}_i$, a plane is fitted by least squares to the reliable support cells in a broad neighborhood $\mathcal{N}^{\mathrm{grade}}_i$ centered on and including $c_i$. Writing $\Delta x_{ij}=x_j-x_i$ and $\Delta y_{ij}=y_j-y_i$ for the planar offsets between cell centers, the plane parameters are obtained as
\begin{equation}
    \begin{aligned}
        (a_i,b_i,d_i)
        ={}&\underset{a,b,d}{\operatorname{argmin}}
        \sum_{j\in\mathcal{N}^{\mathrm{grade}}_i}
        \left[(\hat{z}_j-\hat{z}_i)\right.\\
        &\left.\quad
        -(a\,\Delta x_{ij}+b\,\Delta y_{ij}+d)\right]^2.
    \end{aligned}
    \label{eq:terrain_grade_plane_draft}
\end{equation}
The terrain-grade is the inclination of this plane,
\begin{equation}
    \theta_i^{\mathrm{grade}}
    =\tan^{-1}\!\left(\sqrt{a_i^2+b_i^2}\right).
    \label{eq:terrain_grade_draft}
\end{equation}
It represents the large-scale slope of the underlying terrain rather than an individual step or protrusion. A cell becomes a grade-blocking candidate when $\theta_i^{\mathrm{grade}}$ exceeds the maximum inclination configured for the robot. \Cref{fig:mission_overlay}-A shows the terrain-grade estimate of the mission area, where the difference in slope between the flat road and the inclined hill can be seen. 

Local discontinuities are evaluated relative to the fitted grade. For a cell $c_j$ in a smaller neighborhood $\mathcal{N}^{\mathrm{res}}_i$, the residual height change is
\begin{equation}
    r_{ij}
    = (\hat{z}_j-\hat{z}_i)
    -(a_i\Delta x_{ij}+b_i\Delta y_{ij}).
    \label{eq:terrain_residual_height_draft}
\end{equation}
The residual-slope estimate is then formed from the largest residual that exceeds the minimum height criterion,
\begin{equation}
\begin{aligned}
    \theta_i^{\mathrm{res}}
    &=\max_{\substack{j\in\mathcal{N}^{\mathrm{res}}_i\\r_{ij}>h_{\min}}}
    \tan^{-1}\!\left(
        \frac{r_{ij}}
        {\sqrt{\Delta x_{ij}^{2}+\Delta y_{ij}^{2}}}
    \right),\\
    \theta_i^{\mathrm{res}}&\triangleq 0
    \quad\text{if no }j\text{ satisfies }r_{ij}>h_{\min}.
\end{aligned}
    \label{eq:terrain_residual_slope_draft}
\end{equation}
This quantity identifies local steps and protrusions that cannot be explained by the broader terrain inclination. A residual-blocking candidate is produced only when the residual-slope limit is exceeded, and the height criterion is supported by a minimum number of neighboring cells. As seen in \Cref{fig:mission_overlay}-B, the residual-slope estimate picks up obstacles that break the continuity of the surface, which the robot is unable to overcome.

To simplify collision checking by the planner, the mapper constructs a conservative traversability mask that excludes cells classified as untraversable or having a high geometric $\lambda$ intensity. Small gaps in the raw mask are closed, holes are filled, and obstacle regions are inflated to account for the robot footprint.

The mapper exposes three distinct information layers. First, the free-space mask encodes free, occupied, inflated, and unknown cells after geometric blocking, morphological closing, enclosed-void filling, and robot-footprint inflation. Second, the uncertainty score is a normalized form of the expected reduction $\Delta V_{i,k}$ for the active hazard-size hypothesis. Third, the map of remaining observations combines the free-space class with an uncertainty score. It assigns high demand to reachable cells that still benefit from observation and low or zero demand to sufficiently observed or occupied cells. 

These products preserve the distinction between a cell that is currently traversable, one that contains geometric or hazard evidence, and one that remains insufficiently observed.

\begin{figure*}[t]
    \centering
    
    \includegraphics[width=0.98\linewidth,keepaspectratio]{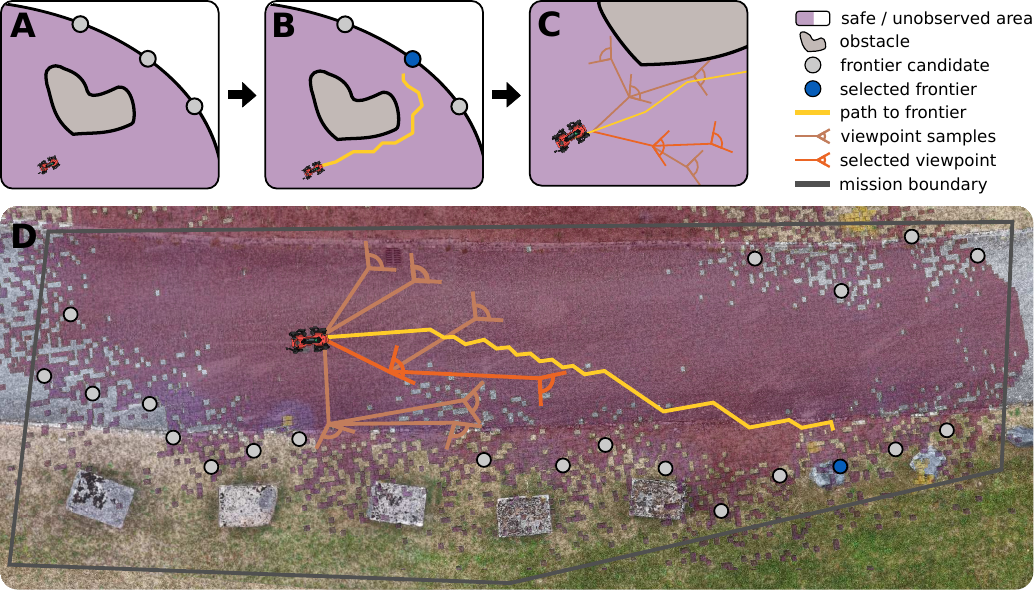}
    \caption{A summary of the steps of the path planning process. \textbf{(A)} Formation of frontier points. \textbf{(B)} Construction of a rough path towards the targeted frontier point. \textbf{(C)} Candidate viewpoints, their respective paths, and the chosen path. \textbf{(D)} Map and planning visuals overlaid on aerial imagery of a mission area.}
    \label{fig:ppr_small_top_down}
\end{figure*}

\subsection{Map-Aware Exploration Planning}
\label{sec:planning_draft}

The exploration planner uses a hierarchical structure. The global planner decides which region of the mission area should be explored next. It detects the boundary between the sufficiently observed and the remaining areas, known as the frontier. The frontier candidates are ranked in terms of proximity to the robot's location and the potential information gain that might be received if the robot visits the respective point. The highest-ranked frontier point is elected as the active objective. The local planner then computes a short informative pose sequence toward this objective. Both levels use the outputs of the mapping framework, including the traversability mask, hazard-uncertainty score, remaining-observation map, and underlying intensity values. Both layers of the planner additionally use the mission boundary supplied before deployment to restrict candidate motions.

\subsubsection{Global Exploration Objective}
\label{sec:global_planning_draft}

The frontier detection is performed using a configured RRT detector, equivalent to Zhao et al.~\cite{zhao_tdle_2023}. 
Frontiers are defined as the cells on the border between sufficiently observed (safe) areas and the remaining areas, shown in purple and white in \Cref{fig:ppr_small_top_down}, respectively.

Raw frontier cells are then merged within a radius into frontier candidates (gray dots in \Cref{fig:ppr_small_top_down}), which are retained only while they lie inside the mission boundary and remain adjacent to insufficiently observed cells ~\cite{zhao_tdle_2023}. 

The frontier candidate ranking uses a simplified exploration-demand score. Each frontier candidate is scored by the mean remaining-observation demand in a local disk around the candidate and, optionally, by the normalized travel distance from the robot. The score is
\begin{equation}
    s = (1-\alpha)G_{\mathrm{area}} - \alpha D_{\mathrm{robot}},
\end{equation}
where $G_{\mathrm{area}}$ is the local exploration-demand gain and $D_{\mathrm{robot}}$ is the saturated robot-to-frontier distance. Setting $\alpha=0$ means the frontier selection is driven primarily by exploration demand, while $\alpha=1$ means the robot prefers exploiting local and accessible frontiers. The frontier candidate that ranks the highest according to this metric is the selected frontier (blue dot in \Cref{fig:ppr_small_top_down}).
To prevent oscillation between different frontiers, a new frontier only replaces the current one when its score exceeds the current one by a configured margin.

An $A ^*$ \cite{hart_formal_1968} path is calculated from the robot to the selected frontier on the lambda field grid. This coarse reference path is shown as a yellow line in \Cref{fig:ppr_small_top_down}.
The search runs on a $3 \times 3$ down-sampled grid, which is sufficient for a coarse reference path.
The search is restricted to a bounding window around the start-goal box, with increasing margins if no valid path is found. 
Through these measures, we reduce the computational burden of the process. 
This coarse reference path only considers regions that are sufficiently observed and deemed safely traversable. The coarse path is not the final executed path; it is a global progress reference used to keep the local informative planner from diverging away from the path towards the selected frontier.

\subsubsection{Local Informative Trajectory Generation}
\label{sec:local_planning_draft}

The local planner generates an informative pose tree using the selected frontier, the coarse reference path, and the map layers provided by the mapper. 

The root of the tree is normally the current robot pose. When the robot is already following a previous path, the tree can be re-rooted at a validated pose ahead of the robot, allowing a smoother transition between the committed and newly generated local paths.

To improve sampling efficiency and encourage progress towards the frontier goal, the planner samples viewpoints locally as well as around the course path. 
Each edge of the viewpoint graph is collision-checked at discrete points along its length against the mission boundary and the traversability map. The geometric risk of each feasible edge is then computed using the lambda field formulation in Eq.~\eqref{eq:geometric_lambda_estimation_draft}.
In \Cref{fig:ppr_small_top_down}, the viewpoint samples are given with the brown lines and expanded around the coarse path (yellow line) towards the selected frontier.

When the pan--tilt unit (PTU) is enabled, the initial sampling is done for the camera viewpoint angle, where the most informative angle that is readily achievable is chosen per candidate. Afterwards, the robot heading angles are sampled, but they are constrained by the PTU pan angle limits as the PTU can only achieve angles $\theta_{pan} \in [-\theta_{limit},\theta_{limit}]$. The robot heading candidate that results in the least amount of change from the current state is then chosen.
 
The OPI-information gain of a feasible edge is evaluated at a small number of poses along the edge. At each pose, a local visibility field is computed from the obstacle mask using the method of Ibrahim et al.~\cite{ibrahim_efficient_2026}. The camera model applies the sensor field of view (FOV), range limit, and minimum hazard-viewing range. Only cells that are within the valid sensing region and not occluded by geometric obstacles contribute to the information score.
For each cell $c_i$ viewed from a candidate pose $\mathbf{x}$, the hazard-information gain is
\begin{equation}
    g_i(\mathbf{x})
    =\nu_i(\mathbf{x})\,k_i(\mathbf{x},R^{*})\,\Delta V_{i,k^*},
    \label{eq:local_information_gain_revised_draft}
\end{equation}
where $R^{*}$ is the operating OPI radius, $\nu_i\in\{0,1\}$ is the geometric visibility indicator, and $k_i\in[0,1]$ accounts for the camera FOV, range, and detectability support. If several poses on an edge can observe the same cell, only the maximum contribution is retained; cells already credited earlier on the branch are not counted again. This produces a marginal information score.

The immediate utility of a feasible local edge is
\begin{equation}
\begin{aligned}
    u={}&w_{\mathrm{info}}G_{\mathrm{haz}}+V_{\mathrm{goal}}
    -w_{\mathrm{motion}}C_{\mathrm{motion}} \\
    &-w_{\mathrm{risk}}C_{\mathrm{risk}}+S_{\mathrm{smooth}}.
\end{aligned}
\label{eq:local_edge_utility_draft}
\end{equation}
Here $G_{\mathrm{haz}}$ is the marginal hazard-information gain of the edge, obtained by aggregating Eq.~\eqref{eq:local_information_gain_revised_draft} over the cells it newly observes,
\begin{equation}
    G_{\mathrm{haz}}
    =\sum_{i\in\mathcal{O}(e)}\ \max_{\mathbf{x}\in\mathcal{P}(e)} g_i(\mathbf{x}),
    \label{eq:hazard_gain_aggregation_draft}
\end{equation}
where $\mathcal{P}(e)$ are the sampled poses along edge $e$ and $\mathcal{O}(e)$ are the cells visible from at least one of them that were not already credited to a parent edge of the same branch. The remaining terms are the following: $V_{\mathrm{goal}}$ rewards progress toward the observation goal while penalizing remaining goal distance, $C_{\mathrm{motion}}$ penalizes path length and yaw change, $C_{\mathrm{risk}}$ is the integrated geometric-intensity risk cost, and $S_{\mathrm{smooth}}$ promotes heading and curvature consistency. The cumulative utility of a pose-tree branch determines the preferred local trajectory, highlighted in orange in \Cref{fig:ppr_small_top_down}.

The selected branch is subsequently checked against the coarse reference path. A detour budget limits how far the local planner may deviate from the coarse reference path. Once accepted, the planner begins following the selected path and retains it until it is completed, invalidated, or replaced during replanning.

\providecommand{\methodnote}[1]{%
    \par\noindent\textcolor{blue}{\textbf{Author note:} #1}\par}
\definecolor{expdraftcolor}{RGB}{0,105,75}
\providecommand{\expdraft}[1]{\textcolor{expdraftcolor}{#1}}

\begin{figure}
    \centering
    \includegraphics[width=1.0\linewidth]{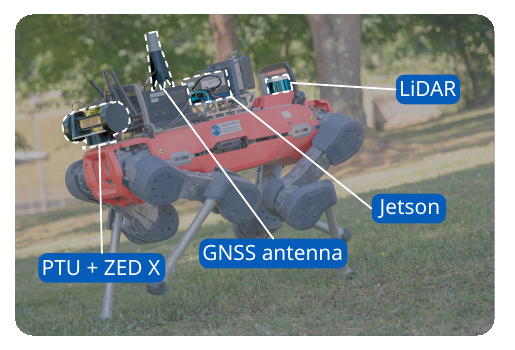}
    \caption{The ANYmal D quadruped deployed for the missions. It is equipped with a Stereolabs ZED X depth camera, an NVIDIA Jetson AGX Orin, and a GNSS unit.}
    \label{fig:robot_dep}
\end{figure}

\begin{figure*}
    \centering
    \includegraphics[width=1.0\linewidth]{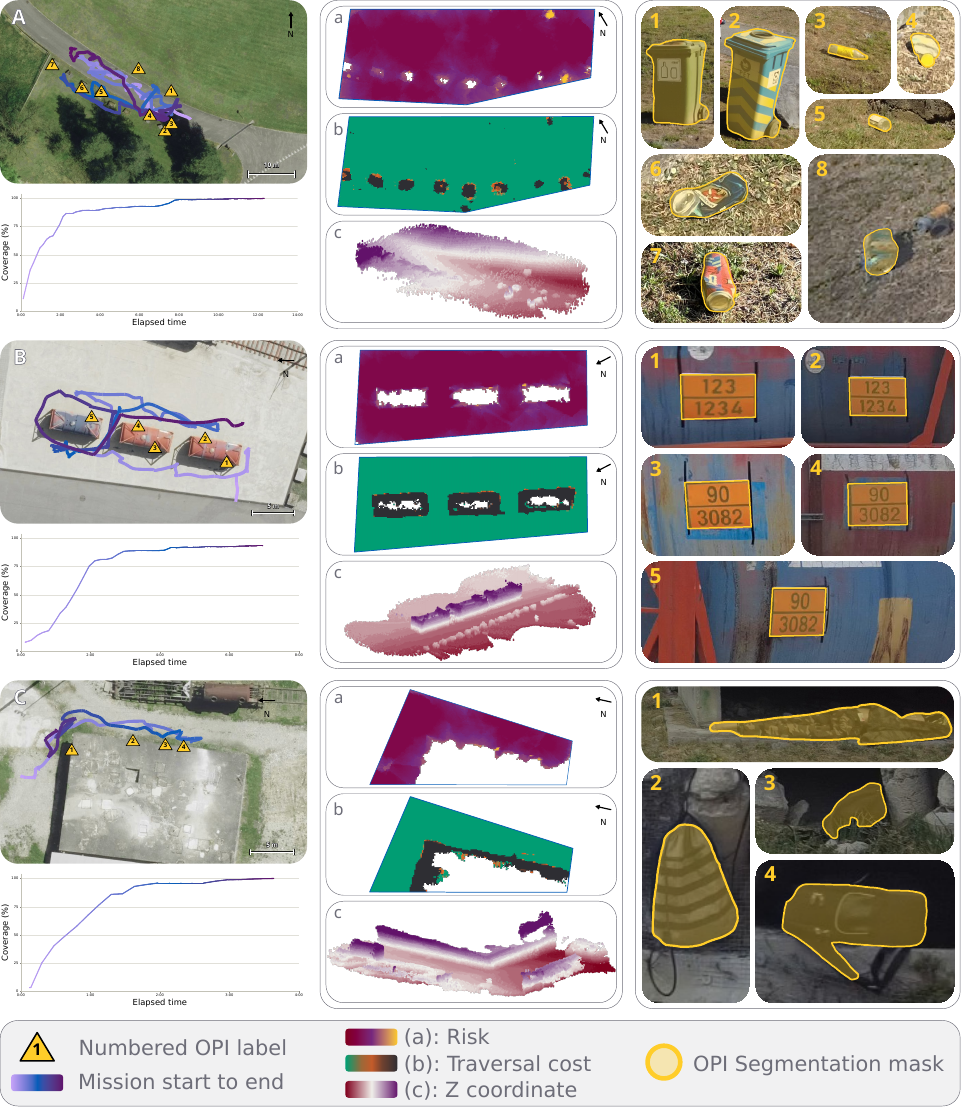}
    \caption{Overview of per-site tasks and primary mission outputs. The leftmost column overlays the path of the robot in the baseline missions, the location of OPIs, and the coverage over time curve. The middle column displays the hazard layer, traversability layer, and accumulated 3-D geometry for each mission. The rightmost column correlates the numbered OPI labels with the actual object, highlighted in yellow. \textbf{(A)} Roadside cleanup; \textbf{(B)} storage tank inspection; \textbf{(C)} disaster response.}
    \label{fig:missions_viz}
\end{figure*}

\begin{table*}[t]
    \centering
    \footnotesize
    \setlength{\tabcolsep}{5pt}
    \begin{tabular}{l | cccccc ccc cc ccc}
        \toprule
        \multirow{2}{*}{Mission}
        & \multicolumn{6}{c}{System metrics}
        & \multicolumn{3}{c}{Coverage}
        & \multicolumn{2}{c}{Detections}
        & \multicolumn{3}{c}{Interventions} \\
        \cmidrule(lr){2-7} \cmidrule(lr){8-10} \cmidrule(lr){11-12} \cmidrule(lr){13-15}
        & Speed & Stand & Path & Dur. & Area & Access. & MCPS & 50/75/90\% & Final & Detections & False Det. & Num. & Duration & No-Int. \\
        & {\scriptsize [\unit[per-mode = symbol]{\metre\per\second}]} & {\scriptsize [\unit{\second}]} & {\scriptsize [\unit{\metre}]} & {\scriptsize [\unit{\second}]} & {\scriptsize [\unit{\square\metre}]} & {\scriptsize [\unit{\square\metre}]}
        & {\scriptsize [\unit[per-mode = symbol]{\square\metre\per\second}]} & {\scriptsize [\unit{\second}]} & {\scriptsize [\%]} & {\scriptsize [\#]} & {\scriptsize [\#]}
        & {\scriptsize [\#]} & {\scriptsize [\unit{\second}]} & {\scriptsize [\%]} \\
        \midrule
        M01 & 0.39 & 198 & 431.5 & 1105 & 319 & 282 & 0.25 & 28/78/180 & 99.98 & 8/8 & 0 & 3 & 8.5 & 98 \\
        M02 & 0.44 & 81 & 216.2 & 495 & 320 & 263 & 0.53 & 38/167/229 & 99.44 & 6/8 & 0 & 6 & 13.7 & 83 \\
        M03 & 0.42 & 89 & 312.7 & 744 & 317 & 275 & 0.36 & 41/107/235 & 99.92 & 8/8 & 0 & 0 & -- & 100 \\
        M04 & 0.41 & 54 & 191.8 & 471 & 318 & 272 & 0.57 & 54/134/154 & 99.60 & 6/8 & 0 & 0 & -- & 100 \\
        M05 & 0.38 & 78 & 239.1 & 636 & 317 & 269 & 0.43 & 46/95/255 & 99.26 & 6/8 & 0 & 0 & -- & 100 \\
        M06 & 0.43 & 59 & 245.7 & 570 & 320 & 287 & 0.52 & 32/74/187 & 99.93 & 4/8 & 0 & 0 & -- & 100 \\
        M07 & 0.44 & 81 & 185.3 & 425 & 278 & 193 & 0.42 & 83/110/243 & 93.42 & 3/5 & 0 & 0 & -- & 100 \\
        M08 & 0.42 & 66 & 156.0 & 375 & 277 & 192 & 0.48 & 95/174/255 & 94.43 & 5/5 & 0 & 0 & -- & 100 \\
        M09 & 0.55 & 32 & 80.5 & 147 & 135 & 63 & 0.43 & 30/71/90 & 100 & 2/4 & 0 & 2 & 10.3 & 86 \\
        M10 & 0.38 & 51 & 84.3 & 221 & 198 & 97 & 0.44 & 32/58/86 & 100 & 3/4 & 2 & 0 & -- & 100 \\
        M11 & 0.47 & 37 & 102.2 & 218 & 199 & 112 & 0.52 & 36/58/83 & 100 & 1/4 & 1 & 0 & -- & 100 \\
        \bottomrule
    \end{tabular}
    \vspace{6pt}
    \caption{Per-mission metrics: system, coverage, detection, and intervention statistics. Speed is path length divided by total recorded duration, including standing time. Standing uses a \qty[per-mode = symbol]{0.05}{\metre\per\second} threshold; Access.\ is the observable subset of the mission area; intervention duration is averaged per takeover.}
    \label{tab:mission_metrics}
\end{table*}

\begin{table}[t]
    \centering
    \footnotesize
    \setlength{\tabcolsep}{5pt}
    \newcommand{\chg}[1]{\textbf{#1}}  
    \resizebox{1.0\linewidth}{!}{%
    \begin{tabular}{l | l l ccc}
        \toprule
        Ablation & Mission & Location & $\alpha$ & Obj.\ radius [m] & PTU \\
        \midrule
        \multirow{3}{*}{\shortstack[l]{Baseline\\\scriptsize(reference config.)}}
            & M03 & road     & 0.0 & 0.2 & off \\
            & M07 & tanks    & 0.0 & 0.2 & off \\
            & M10 & building & 0.0 & 0.2 & off \\
        \midrule
        \multirow{5}{*}{\shortstack[l]{Explor./exploit.\ dial\\\scriptsize(varies $\alpha$)}}
            & M01$^{\dagger}$ & road & \chg{0.3} & 0.2 & \chg{on} \\
            & M05 & road     & \chg{0.5} & 0.2 & off \\
            & M04 & road     & \chg{1.0} & 0.2 & off \\
            & M08 & tanks    & \chg{1.0} & 0.2 & off \\
            & M09 & building & \chg{1.0} & 0.2 & off \\
        \midrule
        \multirow{2}{*}{\shortstack[l]{Active sensing\\\scriptsize(varies PTU)}}
            & M02 & road     & 0.0 & 0.2 & \chg{on} \\
            & M11 & building & 0.0 & 0.2 & \chg{on} \\
        \midrule
        \shortstack[l]{Expected obj.\ size\\\scriptsize(varies neg.\ conf.)}
            & M06 & road     & 0.0 & \chg{1.2} & off \\
        \bottomrule
    \end{tabular}
    }
    \vspace{6pt}
    \caption{Index of all performed ablations. Parameters that differ from the baseline are shown in bold. $^{\dagger}$M01 varies two factors simultaneously.}
    \label{tab:ablation_index}
\end{table}

\section{EXPERIMENTS}
\label{sec:experiments}

We validated the system in a field campaign comprising 11 missions across three sites with distinct task definitions: reporting roadside litter, inspecting a row of storage tanks containing dangerous goods, and responding to a disaster near a partially collapsed building. All missions used the same robot and software modules. \Cref{tab:ablation_index} lists which missions vary each investigated parameter: the exploration-exploitation dial $\alpha$, the active sensing PTU, and the expected object size that drives the confidence of negative detections, and quantifies their effect compared to a baseline. Missions M03, M07, and M10 served as the baseline missions, and the rest account for our ablation study. 

The remainder of this section is organized as follows: \Cref{sec:exp_setup} describes the sites, mission protocol, and evaluation metrics. \Cref{sec:exp_results} inspects mission outputs, reports system-level performance including per-mission motion and intervention statistics, and discusses the observed behavior. Finally, \Cref{sec:exp_ablations} reports observed differences between the investigated configurations.

\subsection{Experimental Setup and Mission Descriptions}
\label{sec:exp_setup}

\Cref{fig:missions_viz} provides a summary of the mission setup for each site. The roadside cleanup area is the largest and most open, \qty{320}{\square\metre} of mostly flat road with scattered roadside objects and small untraversable obstacles. The tank inspection area is \qty{280}{\square\metre}, containing three large cylindrical tanks that serve as both navigation obstacles and large occluders, so any viewpoint covers only a small fraction of the polygon. The disaster response area is the smallest and most constrained at \qty{200}{\square\metre} and is characterized by a partially collapsed building that occupies a large region inside the bounds. \Cref{tab:mission_metrics} lists every mission with its outcome.

The OPIs are selected to be representative of the scenario at hand, but also to span the range of object sizes and detectability. 
For example, the litter reporting task includes trash cans, which are large objects with a high likelihood of detection, aluminum cans, which stand in sharp contrast to the surrounding environment but are small, and transparent PET bottles, which are both small and more difficult to discern from the background.

The pipeline is deployed on an ANYmal D robot \cite{hutter_anymal_2016} with a custom payload: a Stereolabs ZED X RGB-D camera mounted on a custom PTU is used as the perception module; an NVIDIA Jetson AGX Orin \qty{32}{\giga\byte} is used for stereo depth processing and object segmentation; the two processors onboard the robot are used for locomotion, navigation and state estimation; a Global Navigation Satellite System (GNSS) receiver is attached to the robot and supplied through the internet with Real-Time Kinematic (RTK) corrections; an external router is used to manage the robot and payload network connections. \Cref{fig:robot_dep} depicts the robot and highlights the onboard components relevant for our deployments.

Throughout this section, we report coverage relative to the \emph{accessible mission area} $\mathcal{A} \subseteq \Omega$ (reported in \Cref{tab:mission_metrics}), estimated per mission from the final traversability layer: observed traversable cells, plus unobserved cells connected to them through a traversable frontier, which are assumed traversable. Since the mission polygon and observed obstacles nearly enclose $\Omega$ by mission end, this only fills small gaps such as occlusion shadows, and it makes the reported coverage conservative. Coverage is the fraction of $\mathcal{A}$ whose uncertainty score falls below $U_{\min}$ (\Cref{tab:mapping_params}).

Finally, the open-vocabulary prompts differ per scenario and are configured as follows: 
\begin{itemize}
    \item road scenario: \textit{trash can, bottle, can}
    \item tanks scenario: \textit{orange number plate}
    \item building scenario: \textit{human, bag, backpack}.
\end{itemize}

\subsection{Baseline Results}
\label{sec:exp_results}

\subsubsection{Exploration performance}

\Cref{fig:missions_viz} shows the primary output of the system across all three sites, as well as visual references for all emplaced OPIs. The leftmost column displays the coverage over time graph for each mission, and the actual path taken overlaid on satellite imagery. These plots, along with the \textit{``Time to 50/75/90\%''} column of \Cref{tab:mission_metrics}, point to an exploration behavior common to all missions: initially, coverage increases quickly up to \qty{75}{\percent}, after which the rate of coverage decreases steadily until reaching mission end. The time to reach \qty{90}{\percent} coverage is generally equal to or lower than the time needed to cover the last \qty{10}{\percent}. Over the six road missions the median $t_{50}$ is \qty{39.5}{\second}, the median $t_{90}$ is \qty{208}{\second}, and the median total mission time is \qty{603}{\second}.

We attribute this behavior to the per-frontier expected information-gain approach we employ (see \Cref{sec:global_planning_draft}). At the beginning of the mission, frontiers in unoccluded areas of the map are maximally informative, and the distance between them is negligible in terms of traversal time. As the mission progresses, the informativeness of frontiers decreases and the distance between frontiers increases, which corresponds to the lower rate of coverage. The arbitration between frontier informativeness and travel distance is discussed further in \Cref{sec:exp_ablations}.

The robot operated autonomously for \qty{97.6}{\percent} of total mission time; only three missions required intervention. The reasons for interventions are the robot being unable to compute a new path because of perceived false obstacles around the robot's base caused by bad state estimation during high-speed camera panning motions (M01, M02), precautionary safety interventions as the robot was moving directly towards an obstacle (M01, M09), and expediting exploration of the last region of interest due to overheating and low battery (M02).
The unweighted mean of the mission-average speeds is \qty[per-mode = symbol]{0.43}{\metre\per\second} across all sites. Standing time represents \qty{15}{\percent} of total operating time; this aggregate does not identify which processes limit planning frequency.

\subsubsection{Detection performance}

The baseline missions M03, M07, and M10 correctly identify 8/8, 3/5, and 3/4 emplaced OPIs, respectively. Their unweighted mission-average recall is \qty{78.3}{\percent}, while pooling the counts across these missions gives 14/17, or \qty{82.4}{\percent}. False positives occur only in M10, where rubble from the collapsed building is incorrectly identified as a human. Including ablation missions, in the road and tank scenarios, OPIs are detected \qty{80}{\percent} of the time, whereas in the building scenario only \qty{50}{\percent} of the OPI instances are identified correctly. We attribute this drop in performance to the relatively lower contrast between the OPIs in this scenario and their background, as can be clearly seen in \Cref{fig:missions_viz}.
Detection confidences reported by YOLOE are low in absolute terms across all tested classes, which is why the operating threshold is set at \num{0.12}. \Cref{fig:system_overview,fig:overview_viz} display actual detections extracted from missions, along with their confidence. Even in these cases where the OPIs are clearly visible and well-lit, YOLOE reports confidence values less than \num{0.6}.
Our mapping formulation accounts for part of this, as multiple observations of a low-confidence detection accumulate sufficient intensity to register in the hazard layer, and spurious false positives are suppressed. The size-hypothesis mechanism also helps address uncertain detections, which we return to in \Cref{sec:exp_ablations}.

\subsubsection{Mission outputs}

\begin{figure}[t]
    \centering
    \includegraphics[width=\linewidth]{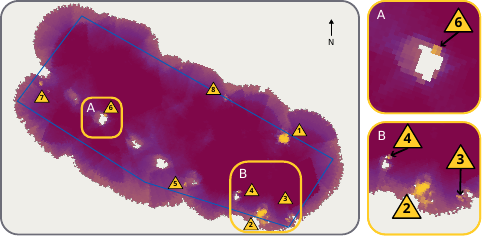}
    \caption{Accumulated risk layer of the M03 mission, shown without clipping to the mission polygon (blue). The numbered OPI labels correspond to the visual references from \Cref{fig:missions_viz}. \textbf{(A)} OPI next to an untraversable granite block; \textbf{(B)} cluster of three OPIs.}
    \label{fig:lambda_raw}
\end{figure}

The (a) panels in the middle column of \Cref{fig:missions_viz} display the final risk layer of the map for each of the baseline missions. In terms of OPI detection, large objects are immediately visible as yellow patches of cells indicating a high $\lambda$ value at that location. In contrast, smaller objects are not readily visible in the map, as they occupy only a few cells, but are still correctly identified as OPIs. \Cref{fig:lambda_raw} highlights these smaller objects in callouts (A) and (B).

Additionally, \Cref{fig:lambda_raw} exemplifies the accumulation mechanism. The map is built from per-ray visibility, and our size hypothesis mechanism assigns lower confidence to observations that are further away from the camera. As a result, the edges of the map hold a higher risk associated with lower confidence, while better-observed regions are categorized as confidently safe. The area of the map that is within mission bounds falls above the well-observed threshold. Empty patches correspond to visibility occluders in the scene, namely the granite blocks on the southern edge of the road.

\Cref{fig:missions_viz}-(A,b), showing the roadside cleanup scenario, best illustrates traversability estimation. A traversable slope is present on the northern edge of the road, and scattered granite blocks prevent traversal on the southern edge of the road. Our system correctly assigns the slope a higher traversal cost, and deems the granite blocks impassable.

\subsection{Ablations}
\label{sec:exp_ablations}

The field ablations varied a single configuration parameter at a time, except for M01, which changed both $\alpha$ and the PTU state. We exclude M01 when assessing either parameter alone. \Cref{tab:ablation_index} lists the parameter settings and associated missions.

\subsubsection{Exploration-Exploitation Dial \texorpdfstring{$\alpha$}{alpha}}

\begin{figure*}[t]
    \centering
    \begin{minipage}[t]{0.32\linewidth}
        \centering
        \includegraphics[width=\linewidth]{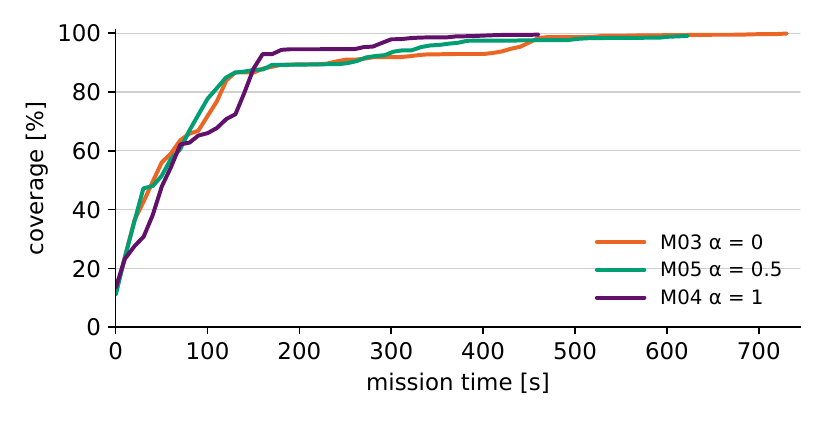}\\
        {\footnotesize (a) Road, $\alpha \in \{0, 0.5, 1\}$}
    \end{minipage}\hfill
    \begin{minipage}[t]{0.32\linewidth}
        \centering
        \includegraphics[width=\linewidth]{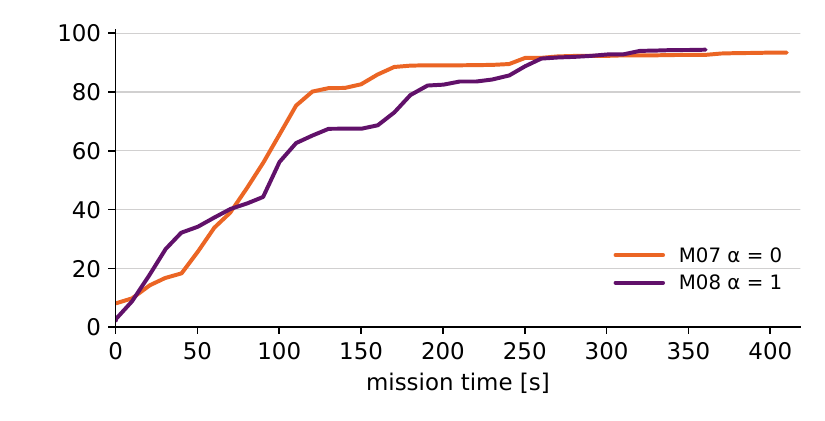}\\
        {\footnotesize (b) Tanks, $\alpha \in \{0, 1\}$}
    \end{minipage}\hfill
    \begin{minipage}[t]{0.32\linewidth}
        \centering
        \includegraphics[width=\linewidth]{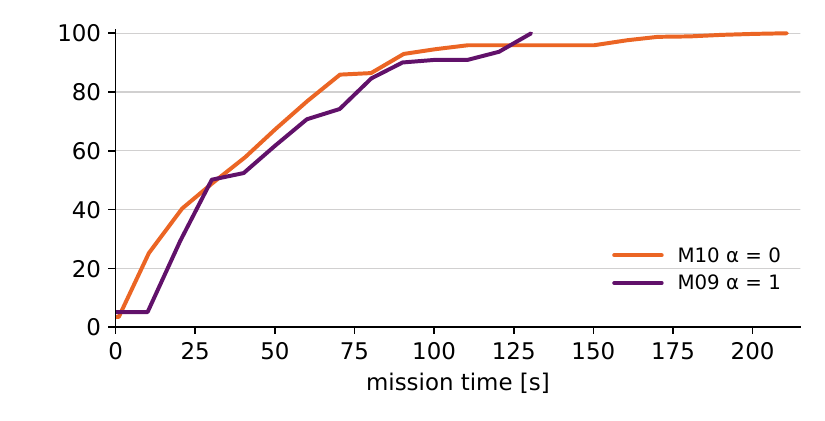}\\
        {\footnotesize (c) Building, $\alpha \in \{0, 1\}$}
    \end{minipage}
    \caption{Effect of the exploration-exploitation dial $\alpha$ on coverage rate. $\alpha=0$ selects frontiers purely by exploration demand; $\alpha=1$ penalizes travel distance and therefore prefers nearby, already-accessible frontiers.}
    \label{fig:ablation_alpha}
\end{figure*}

\Cref{fig:ablation_alpha} compares coverage over time for runs with different values of $\alpha$.
In \Cref{fig:ablation_alpha}-a, for the road scenario, M03 ($\alpha=0$, pure exploration) and M05 ($\alpha=0.5$) are indistinguishable over the first \qty{120}{\second} and reach \qty{85}{\percent} fastest, but M04 ($\alpha=1$, pure exploitation) overtakes them after $\sim$\qty{160}{\second} and completes the mission in \qty{471}{\second} against \qty{636}{\second} (M05) and \qty{744}{\second} (M03), leading to a \qty{37}{\percent} shorter run for M04 than for M03.

In panel (b), for the tank scenario, M07 ($\alpha=0$) reaches \qty{90}{\percent} coverage at \qty{243}{\second}, compared with \qty{255}{\second} for M08 ($\alpha=1$). M08 has a shorter recorded duration, \qty{375}{\second} versus \qty{425}{\second} for M07, so the larger value of $\alpha$ does not improve every coverage milestone.

In panel (c), for the building scenario, M10 ($\alpha=0$) generally has higher coverage through the middle of the plotted run, while M09 ($\alpha=1$) reaches its final coverage sooner. Their recorded durations are \qty{221}{\second} and \qty{147}{\second}, respectively. However, M09 uses a \qty{135}{\square\metre} mission polygon with \qty{63}{\square\metre} accessible area, compared with \qty{198}{\square\metre} and \qty{97}{\square\metre} for M10, and M09 includes two operator interventions. The shorter duration of M09, therefore, cannot be attributed solely to $\alpha$.

Scoring frontiers by information gain alone sends the robot to the most informative frontier anywhere in the map. Late in a mission, this can mean going back and forth between opposite edges of the area.
Early coverage is faster, but the remaining gaps become scattered. They require more traversal to reach, which produces the long tail of the coverage curve. Penalizing distance to the target frontier results in a more structured exploration approach.

\subsubsection{Active Perception with the PTU}

\begin{figure}[t]
    \centering
    \includegraphics[width=1.0\linewidth]{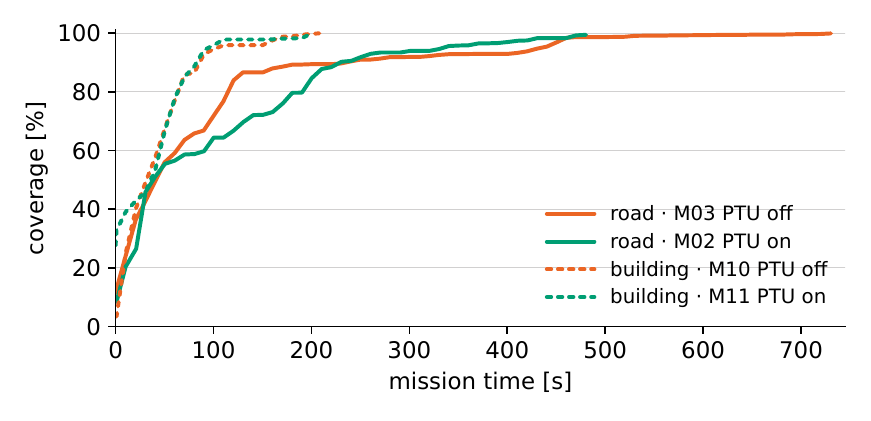}\\
    \caption{Coverage curves with the PTU off and on (road: M03 off vs.\ M02 on; building: M10 off vs.\ M11 on). The runs show no consistent coverage-rate advantage from enabling the PTU.}
    \label{fig:ablation_ptu}
\end{figure}

We compared PTU-on and PTU-off runs at the road and building sites (\Cref{fig:ablation_ptu}).
In the road scene, both runs (M02 and M03) reach \qty{90}{\percent} coverage in similar time and distance covered, namely \qty{230}{\second} and \qty{89}{\metre}, and reach \qty{98}{\percent} coverage at the same time and distance covered. During the last \qty{150}{\second} of the PTU-on mission, the operator intervened in order to expedite the observation of the remaining region of interest, due to low battery and overheating. As such, the mission completion times cannot be directly compared. In the building scene, the two coverage curves remain similar after the first \qty{25}{\second}.

At the road site, PTU-on run M02 detected 6/8 OPIs, compared with 8/8 in PTU-off run M03. At the building site, M11 (PTU on) detected 1/4, compared with 3/4 in M10 (PTU off).

The restricted FOV geometry may contribute to the limited coverage benefit. Full horizontal FOV coverage requires a pan range of $\pm$\ang{125} given the \ang{110} baseline horizontal FOV of the ZED X; however, the additional required payload of the robot limits PTU pan range to $\pm$\ang{40}. This leaves the remaining \ang{170} reachable only by rotating the base. The PTU, therefore, adds viewpoints near the base heading, which may offer limited additional information.

The impact of the PTU also appears in autonomy statistics: at the road site, the two missions where the PTU was enabled recorded nine operator takeovers totaling \qty{1.8}{\minute} out of \qty{26.7}{\minute} of mission time. Two of the interventions can be directly attributed to camera pose estimation errors caused by rapid panning of the camera. These errors inflate and shift the boundaries of obstacles in the environment. The interventions occurred when the obstacle in the map shifted directly onto the robot, blocking further traversal.

Differences in mission area and operator intervention prevent attributing the coverage and detection results to gaze control alone. The restricted pan range, extrinsic calibration burden, and motion artifacts also limit this PTU configuration. Given these confounds and integration limitations, we consider the PTU ablation inconclusive.

\subsubsection{Expected Object Size}

\begin{figure*}[t]
    \centering
    \begin{minipage}[t]{0.32\linewidth}
        \centering
        \includegraphics[width=\linewidth]{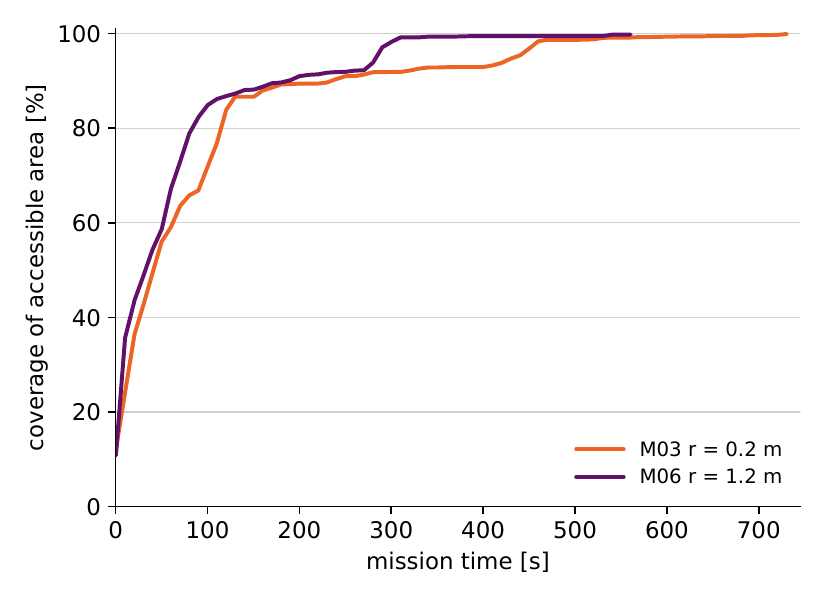}\\
        {\footnotesize (a) Coverage vs.\ time}
    \end{minipage}\hfill
    \begin{minipage}[t]{0.32\linewidth}
        \centering
        \includegraphics[width=\linewidth]{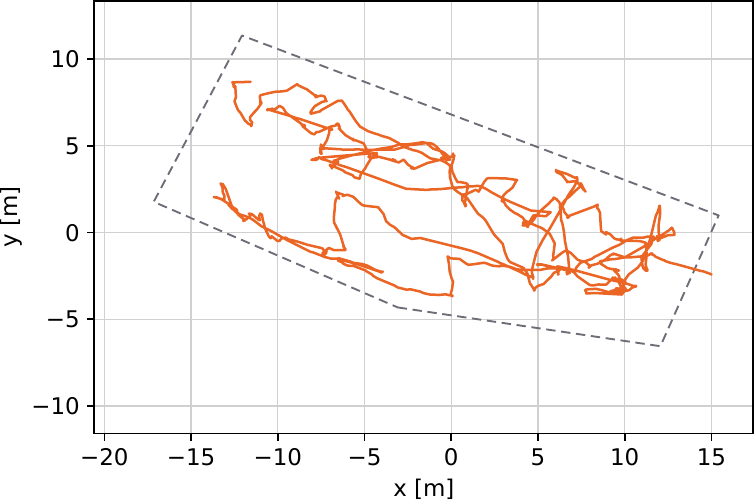}\\
        {\footnotesize (b) M03, $R_{k^*}=0.2\,\mathrm{m}$}
    \end{minipage}\hfill
    \begin{minipage}[t]{0.32\linewidth}
        \centering
        \includegraphics[width=\linewidth]{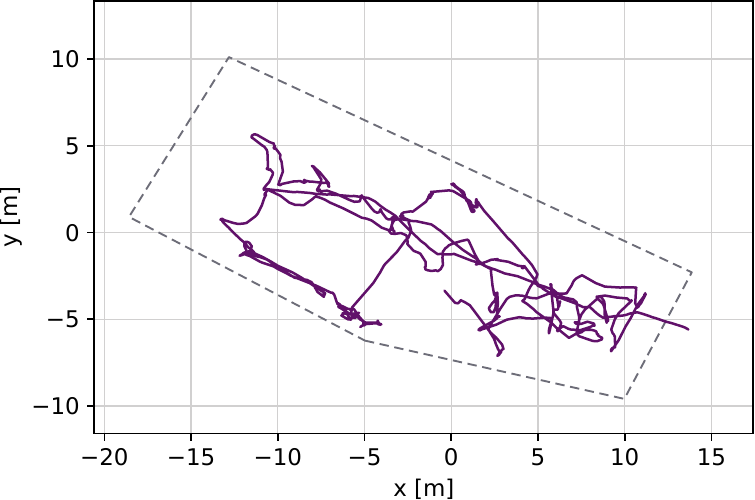}\\
        {\footnotesize (c) M06, $R_{k^*}=1.2\,\mathrm{m}$}
    \end{minipage}
    \caption{Effect of the active hazard-size hypothesis $R_{k^*}$ on search behavior, for two road missions with otherwise identical parameter settings. The size hypothesis sets the range-dependent weighting of negative evidence. Panel (a) shows the corresponding coverage curves.}
    \label{fig:ablation_radius}
\end{figure*}

\begin{figure}
    \centering
    \includegraphics[width=1.0\linewidth]{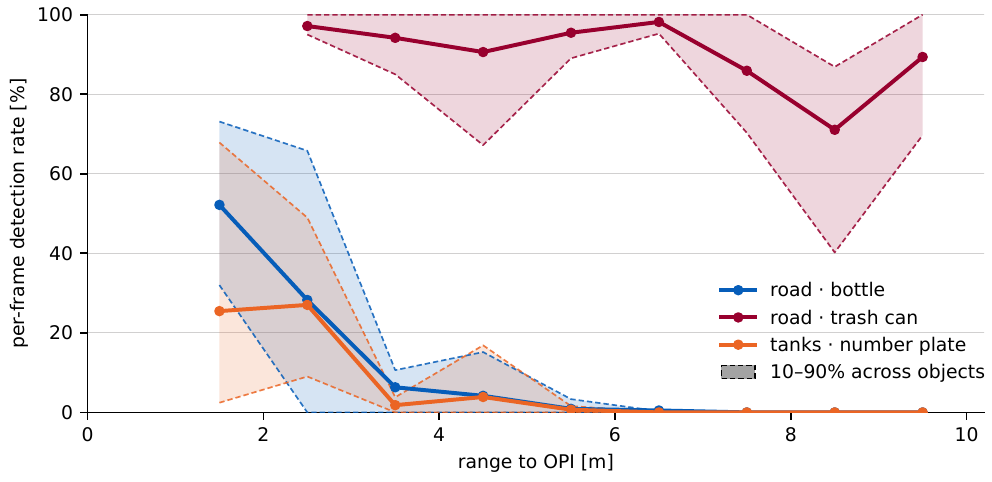}
    \caption{Per-frame detection rate against range to the object, pooled by object class over the missions at the road and tank sites.}
    \label{fig:detection_range}
\end{figure}

The expected object-size hypothesis influences how quickly a cell accumulates observation evidence by setting range-dependent weights for negative updates. We refer to a \emph{non-detection} when a cell lies within the camera view, but no YOLOE detection is associated with it.
The road scenario object sizes span from \qty{1}{\metre} (trash can height) to \qty{15}{\centi\metre} (soda can height), and the plate objects in the tank inspection scenario are \qty{40}{\centi\metre} wide. Both settings use the same maximum visibility range of \qty{6}{\metre}. For $R_{k^*}=\qty{0.2}{\metre}$, Eq.~\eqref{eq:hazard_detectability_ranges_draft} gives $r^{\mathrm{good}}_k\approx\qty{3.61}{\metre}$; the geometric detectability factor decreases beyond this range. For $R_{k^*}=\qty{1.2}{\metre}$, the full-weight range extends beyond the visibility limit, so the factor remains one throughout the sampled range.

Panels (b) and (c) of \Cref{fig:ablation_radius} show the executed trajectories at the road site. At $R_{k^*}=\qty{0.2}{\metre}$ the robot sweeps the area in tight back-and-forth passes. At $R_{k^*}=\qty{1.2}{\metre}$, the larger full-weight range permits stronger negative updates from more distant viewpoints, and the recorded trajectory consists of longer traverses across the polygon. Path length is \qty{21.4}{\percent} shorter and mission time \qty{23}{\percent} lower in the larger-size run.

While coverage is identical across the two runs, the two are not equivalent in outcome. Of the 8 OPIs placed in the road polygon, 8 were detected at $R_{k^*}=\qty{0.2}{\metre}$ against 4 at $R_{k^*}=\qty{1.2}{\metre}$. The two large objects (trash cans) are readily detected in both scenarios, but some smaller litter objects are missing from the $R_{k^*}=\qty{1.2}{\metre}$ run.
Each detection carries a confidence score, which we use to weight positive evidence, but the detector provides no estimate of its miss rate, $P(\text{\textit{no detection}} \mid \text{\textit{OPI present}})$. A non-detection, therefore, cannot be converted into evidence of absence directly. The size hypothesis substitutes a geometric weighting model for this missing quantity. However, meeting the apparent-size threshold does not guarantee detection: transparency, background contrast, illumination, and lens flare also affect whether an object is detected.

\Cref{fig:detection_range} compares the geometric model with measured per-frame detection rates pooled by class. Bottles and number plates are rarely detected beyond \qty{3.5}{\metre}, while trash cans are recovered in over \qty{90}{\percent} of frames out to \qty{6}{\metre} and remain detected at the \qty{10}{\metre} depth ceiling. The observed small-object cutoff is close to $r^{\mathrm{good}}_k\approx\qty{3.61}{\metre}$ at $R_{k^*}=\qty{0.2}{\metre}$, but this is the boundary of the model's full-weight range, not its zero-detectability boundary at approximately \qty{8.02}{\metre}. The numerical proximity, therefore, does not validate the modeled cutoff, and the model permits partial-weight negative evidence at ranges where measured small-object detection is already rare. The absolute detection rate also does not follow size alone. The number plate is physically larger than a bottle, yet peaks at \qty{25}{\percent} compared to the bottle's \qty{52}{\percent} while sharing a similar observed cutoff. The two classes differ in more than size, including background, illumination, and how well each is represented in the detector's training distribution, and our data does not quantify the contribution of these factors.

$R_{k^*}$ therefore controls a trade-off between coverage speed and sensitivity to small objects through the weighting of negative evidence. Choosing $R_{k^*}$ with the smallest mission-relevant object in mind is a heuristic, not a detection guarantee. Closing this gap requires a detector that provides a calibrated estimate of its miss rate, rather than confidence in positive detections alone.

\section{LESSONS LEARNED}
\label{sec:lessons}
Bringing the proposed system to real-world deployment revealed practical challenges spanning architectural decisions, inherent difficulties in operating in potentially hazardous environments, and performance of real-time object detection models. In this section, we discuss these challenges, as well as potential future research directions.

\textbf{PTU integration} on a legged robot platform was constrained by three factors. 
First, the available physical footprint was insufficient for the full range of motion of the camera, limiting its potential coverage advantages. 
Second, estimating the camera-to-base transform during panning motions was a persistent source of error during deployment.
Misalignment between the assumed and true camera pose, particularly in the rotational component, causes compounding measurement errors: a \qty{1}{\degree} offset displaces a detection by \qty{17.5}{\centi\metre} laterally at a range of \qty{10}{\metre}.
Despite the high precision of the unit's motor encoders, calibration drifted in practice because high-frequency oscillations during the stepping phase of the gait led to severe shaking of the PTU and, in one instance, loosened it from the mount entirely. More careful PID tuning reduced the mean oscillation amplitude from \qty{0.68}{\degree} to \qty{0.22}{\degree}. Still, extrinsic misalignment, especially during panning, resulted in phantom obstacles appearing around the robot.
The third factor was that rapid slewing degraded depth measurements by inducing motion blur and occasionally inflated obstacle bounds beyond their actual size.
We therefore plan to reinforce the unit against off-axis skew undetectable by the motor encoders, add a low-level stabilization controller, and drop image sectors that exhibit high motion blur. Complete integration of the PTU, with adequate footprint and a better extrinsic calibration solution, could enable faster and more robust exploration, as well as detection of OPIs placed above the robot's normal vertical range. 

\textbf{Mapping rate} is primarily limited by the number of updates the spatial structure must undergo.
In voxel-based systems, this equates to the number of rays or voxels evaluated per step.
The reported map update rate is \qty{2}{\hertz}. The deployments evaluated here cover mission polygons ranging from \qty{135}{\square\metre} to \qty{320}{\square\metre}; performance over larger areas has not been benchmarked in this campaign.
This rate bounds the exploration speed: the planner treats unmapped areas as unsafe by default, so newly seen ground becomes traversable only once the map update catches up.
The campaign demonstrates exploration at the reported walking speeds with this update rate. Porting to a faster mobile platform, such as a wheeled robot, may require faster map updates.
Specifically, voxel maps can be massively parallelized per-ray and per-voxel, a formulation that would benefit from GPU-based acceleration, thereby enabling mapping rates compatible with higher-speed platforms.

\textbf{Exploration trade-offs} were decided by fixed heuristics.
The mission length during deployments was effectively capped at $30$ minutes by battery life and high ambient temperatures. This tight budget puts direct pressure on the trade-off between exploration speed and risk: moving faster covers more ground but increases the likelihood of catastrophic missed-detection failure modes and overall collision risk. 
Currently, we command a conservative fixed walking speed, which makes the robot unnecessarily slow when re-traversing a known area it is confident in.
The same fixed-heuristic limitation appears in the exploration-versus-exploitation problem. 
As discussed in \Cref{sec:experiments}, our system performs an initial high-coverage pass of the environment, after which it fills in the gaps that were left behind. 
The decision about which areas to leave uncovered during the initial pass accounts only for estimated information gain, not for terrain profile or ease of future access.
A learning-based path planner could arbitrate the two mentioned trade-offs online, taking map uncertainty, OPI detection, and terrain profile into account to directly output base velocity commands. Then, coverage, risk, and traversal time would be implicitly shaped through reward design, rather than tuned via fixed thresholds or hand-specified cost functions.

\textbf{Open vocabulary detectors} were not a research focus of this work, but they directly determined mission success. 
We used YOLOE as an open-vocabulary detector for OPI identification. In practice, its confidence scores vary substantially across classes and are particularly unreliable for classes that are likely underrepresented in its training distribution, requiring per-class rather than global confidence threshold tuning. 
Re-prompting the model for a new set of target classes also carries a practical cost: each prompt change requires re-exporting the model for edge deployment, which takes approximately 30 minutes. 
Furthermore, changes in illumination remained a persistent failure mode independent of prompting. For example, the orange hazard plates produced inconsistent detections depending on whether they were sunlit or in the shadow.

Dense semantic segmentation is a potential alternative direction, particularly given recent advances in self-supervised dense feature extraction. Models such as DINOv3 \cite{simeoni_dinov3_2025} produce high-resolution dense features from a single backbone that enable strong performance across vision tasks, including segmentation, without any task-specific fine-tuning of the backbone itself. These dense per-pixel features are rich enough that lightweight heads can decode them into task-specific outputs. This is particularly attractive for hazard mapping: rather than committing to a single fixed classifier, the same frozen backbone could be paired with several lightweight heads to populate distinct semantic channels on the map. For instance, one head could extend the traversability channel beyond simple geometric analysis with semantic understanding, effectively handling traversal of foliage or loose terrain, while another could populate the OPI-presence channel, which indicates hazard likelihood.

\section{CONCLUSION}
In this work, we presented SAFARI, a framework for autonomous exploration of unknown outdoor scenes in which detected OPIs must both attract observation and repel the platform. Object detections are accumulated as evidence in the same lambda field that carries geometric structure, so that a single representation supplies the traversability, risk and information-gain quantities consumed by a hierarchical frontier and viewpoint planner. We validated the system in 11 missions across three outdoor sites with distinct task definitions, covering \qty{98.7}{\percent} of the accessible mission area on average and operating without operator intervention in eight of the runs.

The PTU-on runs showed no consistent coverage-rate benefit and detected fewer OPIs than the PTU-off runs, although differences in mission area and operator intervention leave the effect of gaze control uncertain. 
The expected-object-size parameter acted as an effective mission-design parameter: the small-object setting recovered all eight road OPIs, including \qty{15}{\centi\metre} cans, while the large-object setting shortened path length by \qty{21}{\percent} and mission time by \qty{23}{\percent} at the same final coverage. 
The exploration-exploitation dial was site dependent: pure exploitation completed the road mission about \qty{37}{\percent} faster than pure exploration, yet neither setting dominated every coverage milestone at the tank and building sites. 
One limitation we found is that the framework treats detector output as evidence, while the detector reports confidence only for presence, not for absence, so non-detections must be weighted by a geometric model rather than a calibrated miss rate. 
Future work will therefore target detectors with calibrated miss-rate estimates and a learned policy that arbitrates coverage, risk, and traversal cost online, replacing the fixed thresholds that currently set the exploration-exploitation dial, walking speed, and the size hypothesis.

\section*{ACKNOWLEDGMENT}
The authors used generative AI to assist with drafting and editing the manuscript. All content was reviewed by the authors, who take full responsibility for it.

\bibliographystyle{IEEEtran}
\bibliography{Bibliography/references}

@inproceedings{hutter_anymal_2016,
    author = {Hutter, Marco and Gehring, Christian and Jud, Dominic and Lauber, Andreas and Bellicoso, C. Dario and Tsounis, Vassilios and Hwangbo, Jemin and Bodie, Karen and Fankhauser, Peter and Bloesch, Michael and Diethelm, Remo and Bachmann, Samuel and Melzer, Amir and Hoepflinger, Mark},
    title = {{ANYmal} - a highly mobile and dynamic quadrupedal robot},
    booktitle = {2016 {IEEE}/{RSJ} {International} {Conference} on {Intelligent} {Robots} and {Systems} ({IROS})},
    pages = {38--44},
    month = oct,
    year = {2016},
    doi = {10.1109/IROS.2016.7758092},
}

@inproceedings{frey_locomotion_2022,
    author = {Frey, Jonas and Hoeller, David and Khattak, Shehryar and Hutter, Marco},
    title = {Locomotion {Policy} {Guided} {Traversability} {Learning} using {Volumetric} {Representations} of {Complex} {Environments}},
    booktitle = {2022 {IEEE}/{RSJ} {International} {Conference} on {Intelligent} {Robots} and {Systems} ({IROS})},
    pages = {5722--5729},
    month = oct,
    year = {2022},
    doi = {10.1109/IROS47612.2022.9982190},
}

@article{hart_formal_1968,
    author = {Hart, Peter E. and Nilsson, Nils J. and Raphael, Bertram},
    title = {A {Formal} {Basis} for the {Heuristic} {Determination} of {Minimum} {Cost} {Paths}},
    journal = {IEEE Transactions on Systems Science and Cybernetics},
    volume = {4},
    number = {2},
    pages = {100--107},
    month = jul,
    year = {1968},
    doi = {10.1109/TSSC.1968.300136},
    issn = {0536-1567},
}

@inproceedings{ibrahim_efficient_2026,
    author = {Ibrahim, Ibrahim and Gillis, Joris and Decré, Wilm and Swevers, Jan},
    title = {An {Efficient} {Solution} to the {2D} {Visibility} {Problem} in {Cartesian} {Grid} {Maps} and its {Application} in {Heuristic} {Path} {Planning}},
    booktitle = {2024 {IEEE} {International} {Conference} on {Robotics} and {Automation} ({ICRA})},
    pages = {7027--7033},
    month = may,
    year = {2024},
    doi = {10.1109/ICRA57147.2024.10611529},
}

@article{fankhauser_probabilistic_2018,
    author = {Fankhauser, Péter and Bloesch, Michael and Hutter, Marco},
    title = {Probabilistic {Terrain} {Mapping} for {Mobile} {Robots} {With} {Uncertain} {Localization}},
    journal = {IEEE Robotics and Automation Letters},
    volume = {3},
    number = {4},
    pages = {3019--3026},
    month = oct,
    year = {2018},
    doi = {10.1109/LRA.2018.2849506},
    issn = {2377-3766},
}

@article{groves_robotic_2021,
    author = {Groves, Keir and Hernandez, Emili and West, Andrew and Wright, Thomas and Lennox, Barry},
    title = {Robotic {Exploration} of an {Unknown} {Nuclear} {Environment} {Using} {Radiation} {Informed} {Autonomous} {Navigation}},
    journal = {Robotics},
    volume = {10},
    number = {2},
    pages = {78},
    month = may,
    year = {2021},
    doi = {10.3390/robotics10020078},
    issn = {2218-6581},
}

@article{west_real-time_2022,
    author = {West, Andrew and Wright, Thomas and Tsitsimpelis, Ioannis and Groves, Keir and Joyce, Malcolm J. and Lennox, Barry},
    title = {Real-{Time} {Avoidance} of {Ionising} {Radiation} {Using} {Layered} {Costmaps} for {Mobile} {Robots}},
    journal = {Frontiers in Robotics and AI},
    volume = {9},
    pages = {862067},
    month = mar,
    year = {2022},
    doi = {10.3389/frobt.2022.862067},
    issn = {2296-9144},
}

@article{nouri_rahmat_abadi_carma_2023,
    author = {Nouri Rahmat Abadi, Bahman and West, Andrew and Peel, Harriet and Nancekievill, Matthew and Ballard, Christopher and Lennox, Barry and Marjanovic, Ognjen and Groves, Keir},
    title = {{CARMA} {II}: {A} ground vehicle for autonomous surveying of alpha, beta and gamma radiation},
    shorttitle = {{CARMA} {II}},
    journal = {Frontiers in Robotics and AI},
    volume = {10},
    pages = {1137750},
    month = mar,
    year = {2023},
    doi = {10.3389/frobt.2023.1137750},
    issn = {2296-9144},
}

@article{robinson_human-robot_2022,
    author = {Robinson, Nicole and Williams, Jason and Howard, David and Tidd, Brendan and Talbot, Fletcher and Wood, Brett and Pitt, Alex and Kottege, Navinda and Kulić, Dana},
    title = {Human-{Robot} {Team} {Performance} {Compared} to {Full} {Robot} {Autonomy} in 16 {Real}-{World} {Search} and {Rescue} {Missions}: {Adaptation} of the {DARPA} {Subterranean} {Challenge}},
    shorttitle = {Human-{Robot} {Team} {Performance} {Compared} to {Full} {Robot} {Autonomy} in 16 {Real}-{World} {Search} and {Rescue} {Missions}},
    journal = {ACM Transactions on Human-Robot Interaction},
    volume = {14},
    number = {1},
    pages = {1--30},
    month = mar,
    year = {2025},
    doi = {10.1145/3665319},
}

@inproceedings{cao_autonomous_2021,
    author = {Cao, Chao and Zhu, Hongbiao and Yang, Fan and Xia, Yukun and Choset, Howie and Oh, Jean and Zhang, Ji},
    title = {Autonomous {Exploration} {Development} {Environment} and the {Planning} {Algorithms}},
    booktitle = {2022 {International} {Conference} on {Robotics} and {Automation} ({ICRA})},
    pages = {8921--8928},
    month = may,
    year = {2022},
    doi = {10.1109/ICRA46639.2022.9812330},
}

@article{zhou_fuel_2020,
    author = {Zhou, Boyu and Zhang, Yichen and Chen, Xinyi and Shen, Shaojie},
    title = {{FUEL}: {Fast} {UAV} {Exploration} using {Incremental} {Frontier} {Structure} and {Hierarchical} {Planning}},
    shorttitle = {{FUEL}},
    journal = {IEEE Robotics and Automation Letters},
    volume = {6},
    number = {2},
    pages = {779--786},
    month = apr,
    year = {2021},
    doi = {10.1109/LRA.2021.3051563},
}

@inproceedings{oleynikova_voxblox_2017,
    author = {Oleynikova, Helen and Taylor, Zachary and Fehr, Marius and Siegwart, Roland and Nieto, Juan},
    title = {Voxblox: {Incremental} {3D} {Euclidean} {Signed} {Distance} {Fields} for {On}-{Board} {MAV} {Planning}},
    shorttitle = {Voxblox},
    booktitle = {2017 {IEEE}/{RSJ} {International} {Conference} on {Intelligent} {Robots} and {Systems} ({IROS})},
    pages = {1366--1373},
    month = sep,
    year = {2017},
    doi = {10.1109/IROS.2017.8202315},
}

@article{dang_graph-based_2020,
    author = {Dang, Tung and Tranzatto, Marco and Khattak, Shehryar and Mascarich, Frank and Alexis, Kostas and Hutter, Marco},
    title = {Graph-based subterranean exploration path planning using aerial and legged robots},
    journal = {Journal of Field Robotics},
    volume = {37},
    number = {8},
    pages = {1363--1388},
    year = {2020},
    doi = {10.1002/rob.21993},
    issn = {1556-4967},
}

@article{liang_hdplanner_2024,
    author = {Liang, Jingsong and Cao, Yuhong and Ma, Yixiao and Zhao, Hanqi and Sartoretti, Guillaume},
    title = {{HDPlanner}: {Advancing} {Autonomous} {Deployments} in {Unknown} {Environments} through {Hierarchical} {Decision} {Networks}},
    shorttitle = {{HDPlanner}},
    journal = {IEEE Robotics and Automation Letters},
    volume = {10},
    number = {1},
    pages = {256--263},
    month = jan,
    year = {2025},
    doi = {10.1109/LRA.2024.3506281},
}

@misc{richter_large-scale_2026,
    author = {Richter, Julia and Tuna, Turcan and Patel, Manthan and Miki, Takahiro and Higgins, Devon and Fox, James and Cadena, Cesar and Diaz, Andres and Hutter, Marco},
    title = {Large-{Scale} {Autonomous} {Gas} {Monitoring} for {Volcanic} {Environments}: {A} {Legged} {Robot} on {Mount} {Etna}},
    shorttitle = {Large-{Scale} {Autonomous} {Gas} {Monitoring} for {Volcanic} {Environments}},
    howpublished = {arXiv preprint},
    month = jan,
    year = {2026},
    doi = {10.48550/arXiv.2601.07362},
    note = {arXiv:2601.07362 [cs.RO], version 2},
    url = {https://arxiv.org/abs/2601.07362v2},
}

@misc{simeoni_dinov3_2025,
    author = {Siméoni, Oriane and Vo, Huy V. and Seitzer, Maximilian and Baldassarre, Federico and Oquab, Maxime and Jose, Cijo and Khalidov, Vasil and Szafraniec, Marc and Yi, Seungeun and Ramamonjisoa, Michaël and Massa, Francisco and Haziza, Daniel and Wehrstedt, Luca and Wang, Jianyuan and Darcet, Timothée and Moutakanni, Théo and Sentana, Leonel and Roberts, Claire and Vedaldi, Andrea and Tolan, Jamie and Brandt, John and Couprie, Camille and Mairal, Julien and Jégou, Hervé and Labatut, Patrick and Bojanowski, Piotr},
    title = {{DINOv3}},
    howpublished = {arXiv preprint},
    month = aug,
    year = {2025},
    doi = {10.48550/arXiv.2508.10104},
    note = {arXiv:2508.10104 [cs.CV], version 1},
    url = {https://arxiv.org/abs/2508.10104v1},
}

@inproceedings{portugal_deploying_2014,
    author = {Portugal, David and Marques, Lino and Armada, Manuel},
    title = {Deploying Field Robots for Humanitarian Demining: Challenges, Requirements and Research Trends},
    shorttitle = {Deploying Field Robots for Humanitarian Demining},
    booktitle = {Mobile {Service} {Robotics}},
    pages = {649--656},
    address = {Poznan, Poland},
    publisher = {World Scientific},
    month = aug,
    year = {2014},
    doi = {10.1142/9789814623353_0075},
    isbn = {978-981-4623-35-3},
}

@article{bajic_uav_2023,
    author = {Bajić, Milan and Potočnik, Božidar},
    title = {{UAV} {Thermal} {Imaging} for {Unexploded} {Ordnance} {Detection} by {Using} {Deep} {Learning}},
    journal = {Remote Sensing},
    volume = {15},
    number = {4},
    pages = {967},
    month = feb,
    year = {2023},
    doi = {10.3390/rs15040967},
    issn = {2072-4292},
}

@article{dasgupta_comrade_2015,
    author = {Dasgupta, Prithviraj and Baca, José and Guruprasad, K. R. and Muñoz-Meléndez, Angélica and Jumadinova, Janyl},
    title = {The {COMRADE} {System} for {Multirobot} {Autonomous} {Landmine} {Detection} in {Postconflict} {Regions}},
    journal = {Journal of Robotics},
    volume = {2015},
    pages = {1--17},
    year = {2015},
    doi = {10.1155/2015/921370},
    issn = {1687-9619},
    note = {Article ID 921370},
}

@article{miles_terrain-aware_2023,
    author = {Miles, Michael J. and Biggie, Harel and Heckman, Christoffer},
    title = {Terrain-aware semantic mapping for cooperative subterranean exploration},
    journal = {Frontiers in Robotics and AI},
    volume = {10},
    pages = {1249586},
    month = oct,
    year = {2023},
    doi = {10.3389/frobt.2023.1249586},
    issn = {2296-9144},
}

@inproceedings{dai_fast_2020,
    author = {Dai, Anna and Papatheodorou, Sotiris and Funk, Nils and Tzoumanikas, Dimos and Leutenegger, Stefan},
    title = {Fast {Frontier}-based {Information}-driven {Autonomous} {Exploration} with an {MAV}},
    booktitle = {2020 {IEEE} {International} {Conference} on {Robotics} and {Automation} ({ICRA})},
    pages = {9570--9576},
    month = may,
    year = {2020},
    doi = {10.1109/ICRA40945.2020.9196707},
}

@book{baudoin_using_2011,
    editor = {Baudoin, Yvan and Habib, Maki K.},
    title = {Using robots in hazardous environments: landmine detection, de-mining and other applications},
    shorttitle = {Using robots in hazardous environments},
    series = {Woodhead publishing in mechanical engineering},
    address = {Cambridge, UK},
    publisher = {Woodhead Publishing},
    year = {2011},
    isbn = {978-1-84569-786-0},
    language = {english},
    doi = {10.1533/9780857090201},
}

@article{border_osprey_2024,
    author = {Border, Rowan and Chebrolu, Nived and Tao, Yifu and Gammell, Jonathan D. and Fallon, Maurice},
    title = {{Osprey}: {Multisession} {Autonomous} {Aerial} {Mapping} with {LiDAR}-based {SLAM} and {Next} {Best} {View} {Planning}},
    shorttitle = {Osprey},
    journal = {IEEE Transactions on Field Robotics},
    volume = {1},
    pages = {113--130},
    year = {2024},
    doi = {10.1109/TFR.2024.3432031},
    issn = {2997-1101},
}

@inproceedings{fan_step_2021,
    author = {Fan, David D. and Otsu, Kyohei and Kubo, Yuki and Dixit, Anushri and Burdick, Joel and Agha-Mohammadi, Ali-Akbar},
    title = {{STEP}: {Stochastic} {Traversability} {Evaluation} and {Planning} for {Risk}-{Aware} {Off}-road {Navigation}},
    shorttitle = {{STEP}},
    booktitle = {Robotics: {Science} and {Systems} ({RSS})},
    month = jul,
    year = {2021},
    doi = {10.15607/RSS.2021.XVII.021},
}

@inproceedings{frey_fast_2023,
    author = {Frey, Jonas and Mattamala, Matias and Chebrolu, Nived and Cadena, Cesar and Fallon, Maurice and Hutter, Marco},
    title = {Fast {Traversability} {Estimation} for {Wild} {Visual} {Navigation}},
    booktitle = {Robotics: {Science} and {Systems} ({RSS})},
    month = jul,
    year = {2023},
    doi = {10.15607/RSS.2023.XIX.054},
}

@inproceedings{wermelinger_navigation_2016,
    author = {Wermelinger, Martin and Fankhauser, Péter and Diethelm, Remo and Krüsi, Philipp and Siegwart, Roland and Hutter, Marco},
    title = {Navigation planning for legged robots in challenging terrain},
    booktitle = {2016 {IEEE}/{RSJ} {International} {Conference} on {Intelligent} {Robots} and {Systems} ({IROS})},
    pages = {1184--1189},
    month = oct,
    year = {2016},
    doi = {10.1109/IROS.2016.7759199},
}

@inproceedings{isler_information_2016,
    author = {Isler, Stefan and Sabzevari, Reza and Delmerico, Jeffrey and Scaramuzza, Davide},
    title = {An information gain formulation for active volumetric {3D} reconstruction},
    booktitle = {2016 {IEEE} {International} {Conference} on {Robotics} and {Automation} ({ICRA})},
    pages = {3477--3484},
    month = may,
    year = {2016},
    doi = {10.1109/ICRA.2016.7487527},
}

@article{border_surface_2024,
    author = {Border, Rowan and Gammell, Jonathan D.},
    title = {The {Surface} {Edge} {Explorer} ({SEE}): {A} measurement-direct approach to next best view planning},
    shorttitle = {The {Surface} {Edge} {Explorer} ({SEE})},
    journal = {The International Journal of Robotics Research},
    volume = {43},
    number = {10},
    pages = {1506--1532},
    month = sep,
    year = {2024},
    doi = {10.1177/02783649241230098},
    issn = {0278-3649},
}

@misc{tankasala_beyond_2024,
    author = {Tankasala, Srinath and Martín-Martín, Roberto and Pryor, Mitch},
    title = {Beyond {Shortsighted} {Navigation}: {Merging} {Best} {View} {Trajectory} {Planning} with {Robot} {Navigation}},
    shorttitle = {Beyond {Shortsighted} {Navigation}},
    howpublished = {arXiv preprint},
    month = aug,
    year = {2024},
    doi = {10.48550/arXiv.2408.12513},
    note = {arXiv:2408.12513 [cs.RO], version 1},
    url = {https://arxiv.org/abs/2408.12513v1},
}

@inproceedings{zhao_tdle_2023,
    author = {Zhao, Xuyang and Yu, Chengpu and Xu, Erpei and Liu, Yixuan},
    title = {{TDLE}: 2-{D} {LiDAR} {Exploration} {With} {Hierarchical} {Planning} {Using} {Regional} {Division}},
    shorttitle = {{TDLE}},
    booktitle = {2023 {IEEE} 19th {International} {Conference} on {Automation} {Science} and {Engineering} ({CASE})},
    pages = {1--6},
    month = aug,
    year = {2023},
    doi = {10.1109/CASE56687.2023.10260441},
}

@inproceedings{randriamiarintsoa_risk-aware_2023,
    author = {Randriamiarintsoa, Elie and Laconte, Johann and Thuilot, Benoit and Aufrère, Romuald},
    title = {Risk-{Aware} {Navigation} for {Mobile} {Robots} in {Unknown} {3D} {Environments}},
    booktitle = {2023 {IEEE} 26th {International} {Conference} on {Intelligent} {Transportation} {Systems} ({ITSC})},
    pages = {1949--1954},
    month = sep,
    year = {2023},
    doi = {10.1109/ITSC57777.2023.10420846},
}

@article{laconte_novel_2021,
    author = {Laconte, Johann and Kasmi, Abderrahim and Pomerleau, François and Chapuis, Roland and Malaterre, Laurent and Debain, Christophe and Aufrère, Romuald},
    title = {A {Novel} {Occupancy} {Mapping} {Framework} for {Risk}-{Aware} {Path} {Planning} in {Unstructured} {Environments}},
    journal = {Sensors},
    volume = {21},
    number = {22},
    pages = {7562},
    month = nov,
    year = {2021},
    doi = {10.3390/s21227562},
    issn = {1424-8220},
}

@article{matez-bandera_voxeland_2024,
    author = {Matez-Bandera, Jose-Luis and Ojeda, Pepe and Monroy, Javier and Gonzalez-Jimenez, Javier and Ruiz-Sarmiento, Jose-Raul},
    title = {Voxeland: {Probabilistic} {Instance}-{Aware} {Semantic} {Mapping} with {Evidence}-based {Uncertainty} {Quantification}},
    shorttitle = {Voxeland},
    journal = {Robotics and Autonomous Systems},
    volume = {205},
    pages = {105668},
    year = {2026},
    doi = {10.1016/j.robot.2026.105668},
}

@article{nagami_vista_2025,
    author = {Nagami, Keiko and Chen, Timothy and Yu, Javier and Shorinwa, Ola and Adang, Maximilian and Dougherty, Carlyn and Cristofalo, Eric and Schwager, Mac},
    title = {{VISTA}: {Open}-{Vocabulary}, {Task}-{Relevant} {Robot} {Exploration} with {Online} {Semantic} {Gaussian} {Splatting}},
    shorttitle = {{VISTA}},
    journal = {IEEE Robotics and Automation Letters},
    volume = {11},
    number = {3},
    pages = {3150--3157},
    month = mar,
    year = {2026},
    doi = {10.1109/LRA.2026.3653276},
}

@article{amatucci_vero_2024,
    author = {Amatucci, Lorenzo and Turrisi, Giulio and Bratta, Angelo and Barasuol, Victor and Semini, Claudio},
    title = {{VERO}: {A} vacuum-cleaner-equipped quadruped robot for efficient litter removal},
    shorttitle = {{VERO}},
    journal = {Journal of Field Robotics},
    volume = {41},
    number = {6},
    pages = {1829--1842},
    year = {2024},
    doi = {10.1002/rob.22350},
    issn = {1556-4967},
}

@article{lima_under-canopy_2024,
    author = {de Lima, Lucas Carvalho and Lawrance, Nicholas and Khosoussi, Kasra and Borges, Paulo and Brünig, Michael},
    title = {Under-{Canopy} {Navigation} using {Aerial} {Lidar} {Maps}},
    journal = {IEEE Robotics and Automation Letters},
    volume = {9},
    number = {8},
    pages = {7031--7038},
    month = aug,
    year = {2024},
    doi = {10.1109/LRA.2024.3417115},
}

@article{tranzatto_team_2022,
    author = {Tranzatto, Marco and Dharmadhikari, Mihir and Bernreiter, Lukas and Camurri, Marco and Khattak, Shehryar and Mascarich, Frank and Pfreundschuh, Patrick and Wisth, David and Zimmermann, Samuel and Kulkarni, Mihir and Reijgwart, Victor and Casseau, Benoit and Homberger, Timon and De Petris, Paolo and Ott, Lionel and Tubby, Wayne and Waibel, Gabriel and Nguyen, Huan and Cadena, Cesar and Buchanan, Russell and Wellhausen, Lorenz and Khedekar, Nikhil and Andersson, Olov and Zhang, Lintong and Miki, Takahiro and Dang, Tung and Mattamala, Matias and Montenegro, Markus and Meyer, Konrad and Wu, Xiangyu and Briod, Adrien and Mueller, Mark and Fallon, Maurice and Siegwart, Roland and Hutter, Marco and Alexis, Kostas},
    title = {Team {CERBERUS} {Wins} the {DARPA} {Subterranean} {Challenge}: {Technical} {Overview} and {Lessons} {Learned}},
    shorttitle = {Team {CERBERUS} {Wins} the {DARPA} {Subterranean} {Challenge}},
    journal = {Field Robotics},
    volume = {4},
    pages = {249--312},
    year = {2024},
    doi = {10.55417/fr.2024009},
}

@article{malczyk_semantically-driven_2025,
    author = {Malczyk, Grzegorz and Kulkarni, Mihir and Alexis, Kostas},
    title = {Semantically-driven {Deep} {Reinforcement} {Learning} for {Inspection} {Path} {Planning}},
    journal = {IEEE Robotics and Automation Letters},
    volume = {10},
    number = {7},
    pages = {7206--7213},
    month = jul,
    year = {2025},
    doi = {10.1109/LRA.2025.3575331},
}

@misc{nguyen_semantic-aware_2026,
    author = {Nguyen, Duc-Thien and Do, Ngoc Minh and HoangVan, Xiem and Canh, Thanh Nguyen},
    title = {Semantic-{Aware} {Autonomous} {Exploration} for {UAVs} in {Unknown} {Indoor} {Environments}},
    howpublished = {arXiv preprint},
    month = jun,
    year = {2026},
    doi = {10.48550/arXiv.2606.22670},
    note = {arXiv:2606.22670 [cs.RO], version 1},
    url = {https://arxiv.org/abs/2606.22670v1},
}

@article{deng_see-csom_2024,
    author = {Deng, Yinan and Wang, Meiling and Yang, Yi and Wang, Danwei and Yue, Yufeng},
    title = {{SEE}-{CSOM}: {Sharp}-{Edged} and {Efficient} {Continuous} {Semantic} {Occupancy} {Mapping} for {Mobile} {Robots}},
    shorttitle = {{SEE}-{CSOM}},
    journal = {IEEE Transactions on Industrial Electronics},
    volume = {71},
    number = {2},
    pages = {1718--1728},
    month = feb,
    year = {2024},
    doi = {10.1109/TIE.2023.3262857},
    issn = {1557-9948},
}

@article{kim_safe_2025,
    author = {Kim, Dabin and Han, Youngsoo and Kim, H. Jin},
    title = {Safe {Reactive} {Control} for {Autonomous} {Navigation} in {Unknown} {Environments} via {Probabilistic} {Maps}},
    journal = {IEEE Access},
    volume = {13},
    pages = {171850--171862},
    year = {2025},
    doi = {10.1109/ACCESS.2025.3610851},
    issn = {2169-3536},
}

@misc{razavi_online_2025,
    author = {Razavi, Emad and Bratta, Angelo and Soares, João Carlos Virgolino and Recchiuto, Carmine and Semini, Claudio},
    title = {Online {Object}-{Level} {Semantic} {Mapping} for {Quadrupeds} in {Real}-{World} {Environments}},
    howpublished = {arXiv preprint},
    month = oct,
    year = {2025},
    doi = {10.48550/arXiv.2510.18776},
    note = {arXiv:2510.18776 [cs.RO], version 1},
    url = {https://arxiv.org/abs/2510.18776v1},
}

@inproceedings{canh_object-oriented_2024,
    author = {Canh, Thanh Nguyen and Elibol, Armagan and Chong, Nak Young and HoangVan, Xiem},
    title = {Object-{Oriented} {Semantic} {Mapping} for {Reliable} {UAVs} {Navigation}},
    booktitle = {2023 12th {International} {Conference} on {Control}, {Automation} and {Information} {Sciences} ({ICCAIS})},
    pages = {139--144},
    month = nov,
    year = {2023},
    doi = {10.1109/ICCAIS59597.2023.10382351},
}

@article{gan_multitask_2023,
    author = {Gan, Lu and Kim, Youngji and Grizzle, Jessy W. and Walls, Jeffrey M. and Kim, Ayoung and Eustice, Ryan M. and Ghaffari, Maani},
    title = {Multitask {Learning} for {Scalable} and {Dense} {Multilayer} {Bayesian} {Map} {Inference}},
    journal = {IEEE Transactions on Robotics},
    volume = {39},
    number = {1},
    pages = {699--717},
    month = feb,
    year = {2023},
    doi = {10.1109/TRO.2022.3197106},
    issn = {1552-3098},
}

@misc{qu_ippon_2024,
    author = {Qu, Kaixian and Tan, Jie and Zhang, Tingnan and Xia, Fei and Cadena, Cesar and Hutter, Marco},
    title = {{IPPON}: {Common} {Sense} {Guided} {Informative} {Path} {Planning} for {Object} {Goal} {Navigation}},
    shorttitle = {{IPPON}},
    howpublished = {arXiv preprint},
    month = oct,
    year = {2024},
    doi = {10.48550/arXiv.2410.19697},
    note = {arXiv:2410.19697 [cs.RO], version 1},
    url = {https://arxiv.org/abs/2410.19697v1},
}

@inproceedings{overbye_g-vom_2021,
    author = {Overbye, Timothy and Saripalli, Srikanth},
    title = {G-{VOM}: {A} {GPU} {Accelerated} {Voxel} {Off}-{Road} {Mapping} {System}},
    shorttitle = {G-{VOM}},
    booktitle = {2022 {IEEE} {Intelligent} {Vehicles} {Symposium} ({IV})},
    pages = {1480--1486},
    month = jun,
    year = {2022},
    doi = {10.1109/IV51971.2022.9827107},
}

@article{ruetz_foresttrav_2024,
    author = {Ruetz, Fabio A. and Lawrance, Nicholas and Hernández, Emili and Borges, Paulo V. K. and Peynot, Thierry},
    title = {{ForestTrav}: {3D} {LiDAR}-{Only} {Forest} {Traversability} {Estimation} for {Autonomous} {Ground} {Vehicles}},
    shorttitle = {{ForestTrav}},
    journal = {IEEE Access},
    volume = {12},
    pages = {37192--37206},
    year = {2024},
    doi = {10.1109/ACCESS.2024.3373004},
}

@inproceedings{papatheodorou_finding_2023,
    author = {Papatheodorou, Sotiris and Funk, Nils and Tzoumanikas, Dimos and Choi, Christopher and Xu, Binbin and Leutenegger, Stefan},
    title = {Finding {Things} in the {Unknown}: {Semantic} {Object}-{Centric} {Exploration} with an {MAV}},
    shorttitle = {Finding {Things} in the {Unknown}},
    booktitle = {2023 {IEEE} {International} {Conference} on {Robotics} and {Automation} ({ICRA})},
    pages = {3339--3345},
    month = may,
    year = {2023},
    doi = {10.1109/ICRA48891.2023.10160490},
}

@article{dumencic_experimental_2025,
    author = {Dumenčić, Stella and Lanča, Luka and Jakac, Karlo and Ivić, Stefan},
    title = {Experimental validation of {UAV} search and detection system in real wilderness environment},
    journal = {Drones},
    volume = {9},
    number = {7},
    pages = {473},
    month = jul,
    year = {2025},
    doi = {10.3390/drones9070473},
    issn = {2504-446X},
}

@inproceedings{kim_evidential_2024,
    author = {Kim, Junyoung and Seo, Junwon and Min, Jihong},
    title = {Evidential {Semantic} {Mapping} in {Off}-road {Environments} with {Uncertainty}-aware {Bayesian} {Kernel} {Inference}},
    booktitle = {2024 {IEEE}/{RSJ} {International} {Conference} on {Intelligent} {Robots} and {Systems} ({IROS})},
    pages = {1420--1427},
    month = oct,
    year = {2024},
    doi = {10.1109/IROS58592.2024.10802766},
}

@inproceedings{laconte_dynamic_2021,
    author = {Laconte, Johann and Randriamiarintsoa, Elie and Kasmi, Abderrahim and Pomerleau, François and Chapuis, Roland and Debain, Christophe and Aufrère, Romuald},
    title = {Dynamic {Lambda}-{Field}: {A} {Counterpart} of the {Bayesian} {Occupancy} {Grid} for {Risk} {Assessment} in {Dynamic} {Environments}},
    shorttitle = {Dynamic {Lambda}-{Field}},
    booktitle = {2021 {IEEE}/{RSJ} {International} {Conference} on {Intelligent} {Robots} and {Systems} ({IROS})},
    pages = {4846--4853},
    month = sep,
    year = {2021},
    doi = {10.1109/IROS51168.2021.9636804},
}

@misc{zhao_decision-driven_2026,
    author = {Zhao, Guoyang and Li, Yudong and Qi, Weiqing and Zhang, Kai and Liu, Bonan and Chen, Kai and Li, Haoang and Ma, Jun},
    title = {Decision-{Driven} {Semantic} {Object} {Exploration} for {Legged} {Robots} via {Confidence}-{Calibrated} {Perception} and {Topological} {Subgoal} {Selection}},
    howpublished = {arXiv preprint},
    month = mar,
    year = {2026},
    doi = {10.48550/arXiv.2509.20739},
    note = {arXiv:2509.20739 [cs.RO], version 2},
    url = {https://arxiv.org/abs/2509.20739v2},
}

@article{mattamala_building_2025,
    author = {Mattamala, Matías and Chebrolu, Nived and Frey, Jonas and Freißmuth, Leonard and Oh, Haedam and Casseau, Benoit and Hutter, Marco and Fallon, Maurice},
    title = {Building {Forest} {Inventories} with {Autonomous} {Legged} {Robots} -- {System}, {Lessons}, and {Challenges} {Ahead}},
    journal = {IEEE Transactions on Field Robotics},
    volume = {2},
    pages = {418--436},
    year = {2025},
    doi = {10.1109/TFR.2025.3583972},
    issn = {2997-1101},
}

@article{kraft_autonomous_2021,
    author = {Kraft, Marek and Piechocki, Mateusz and Ptak, Bartosz and Walas, Krzysztof},
    title = {Autonomous, {Onboard} {Vision}-{Based} {Trash} and {Litter} {Detection} in {Low} {Altitude} {Aerial} {Images} {Collected} by an {Unmanned} {Aerial} {Vehicle}},
    journal = {Remote Sensing},
    volume = {13},
    number = {5},
    pages = {965},
    month = mar,
    year = {2021},
    doi = {10.3390/rs13050965},
    issn = {2072-4292},
}

@article{zhang_autonomous_2024,
    author = {Zhang, Bo and Chen, Weili and Xu, Chaoming and Qiu, Jinshi and Chen, Shiyu},
    title = {Autonomous {Vehicles} {Traversability} {Mapping} {Fusing} {Semantic}–{Geometric} in {Off}-{Road} {Navigation}},
    journal = {Drones},
    volume = {8},
    number = {9},
    pages = {496},
    month = sep,
    year = {2024},
    doi = {10.3390/drones8090496},
    issn = {2504-446X},
}

@article{miller_air-ground_2024,
    author = {Miller, Ian D. and Cladera, Fernando and Smith, Trey and Taylor, Camillo Jose and Kumar, Vijay},
    title = {Air-{Ground} {Collaboration} with {SPOMP}: {Semantic} {Panoramic} {Online} {Mapping} and {Planning}},
    shorttitle = {Air-{Ground} {Collaboration} with {SPOMP}},
    journal = {IEEE Transactions on Field Robotics},
    volume = {1},
    pages = {93--112},
    year = {2024},
    doi = {10.1109/TFR.2024.3424748},
    issn = {2997-1101},
}

@inproceedings{wang_actively_2020,
    author = {Wang, Yiduo and Ramezani, Milad and Fallon, Maurice},
    title = {Actively {Mapping} {Industrial} {Structures} with {Information} {Gain}-{Based} {Planning} on a {Quadruped} {Robot}},
    booktitle = {2020 {IEEE} {International} {Conference} on {Robotics} and {Automation} ({ICRA})},
    pages = {8609--8615},
    month = may,
    year = {2020},
    doi = {10.1109/ICRA40945.2020.9197153},
}

@inproceedings{liu_active_2023,
    author = {Liu, Xu and Prabhu, Ankit and Cladera, Fernando and Miller, Ian D. and Zhou, Lifeng and Taylor, Camillo J. and Kumar, Vijay},
    title = {Active {Metric}-{Semantic} {Mapping} by {Multiple} {Aerial} {Robots}},
    booktitle = {2023 {IEEE} {International} {Conference} on {Robotics} and {Automation} ({ICRA})},
    pages = {3282--3288},
    month = may,
    year = {2023},
    doi = {10.1109/ICRA48891.2023.10161564},
}

@article{ulloa_bio-inspired_2026,
    author = {Ulloa, Christyan Cruz and Mateos, Alicia Prieto and del Cerro, Jaime and Barrientos, Antonio},
    title = {Bio-inspired risk-aware adaptive navigation for quadruped robots in search and rescue scenarios},
    journal = {Biomimetic Intelligence and Robotics},
    pages = {100330},
    month = mar,
    year = {2026},
    doi = {10.1016/j.birob.2026.100330},
    issn = {2667-3797},
}

@inproceedings{wang_yoloe_2025,
    author = {Wang, Ao and Liu, Lihao and Chen, Hui and Lin, Zijia and Han, Jungong and Ding, Guiguang},
    title = {{YOLOE}: {Real}-{Time} {Seeing} {Anything}},
    shorttitle = {{YOLOE}},
    booktitle = {2025 {IEEE}/{CVF} {International} {Conference} on {Computer} {Vision} ({ICCV})},
    pages = {24591--24602},
    month = oct,
    year = {2025},
    doi = {10.1109/ICCV51701.2025.02280},
}

\ifanonymous
\else
\begin{IEEEbiography}[{\includegraphics[width=1in,height=1.25in,clip,keepaspectratio]{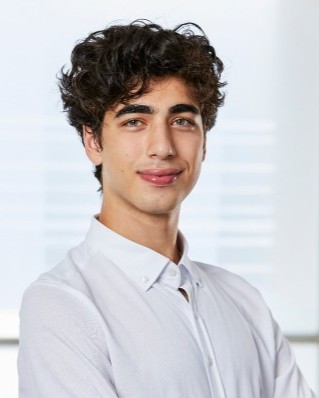}}]
    {\textsc{Ugur Kafkasyali}}~received the B.Sc. degree in mechanical engineering from the Karlsruhe Institute of Technology (KIT), Karlsruhe, Germany, and the M.Sc. degree in mechanical engineering from ETH Zürich, Zürich, Switzerland. During his master's studies, he conducted research at the Robotic Systems Lab (RSL), focusing on autonomous exploration and robotic mapping in unstructured environments. 
    
\end{IEEEbiography}

\begin{IEEEbiography}[{\includegraphics[width=1in,height=1.25in,clip,keepaspectratio]{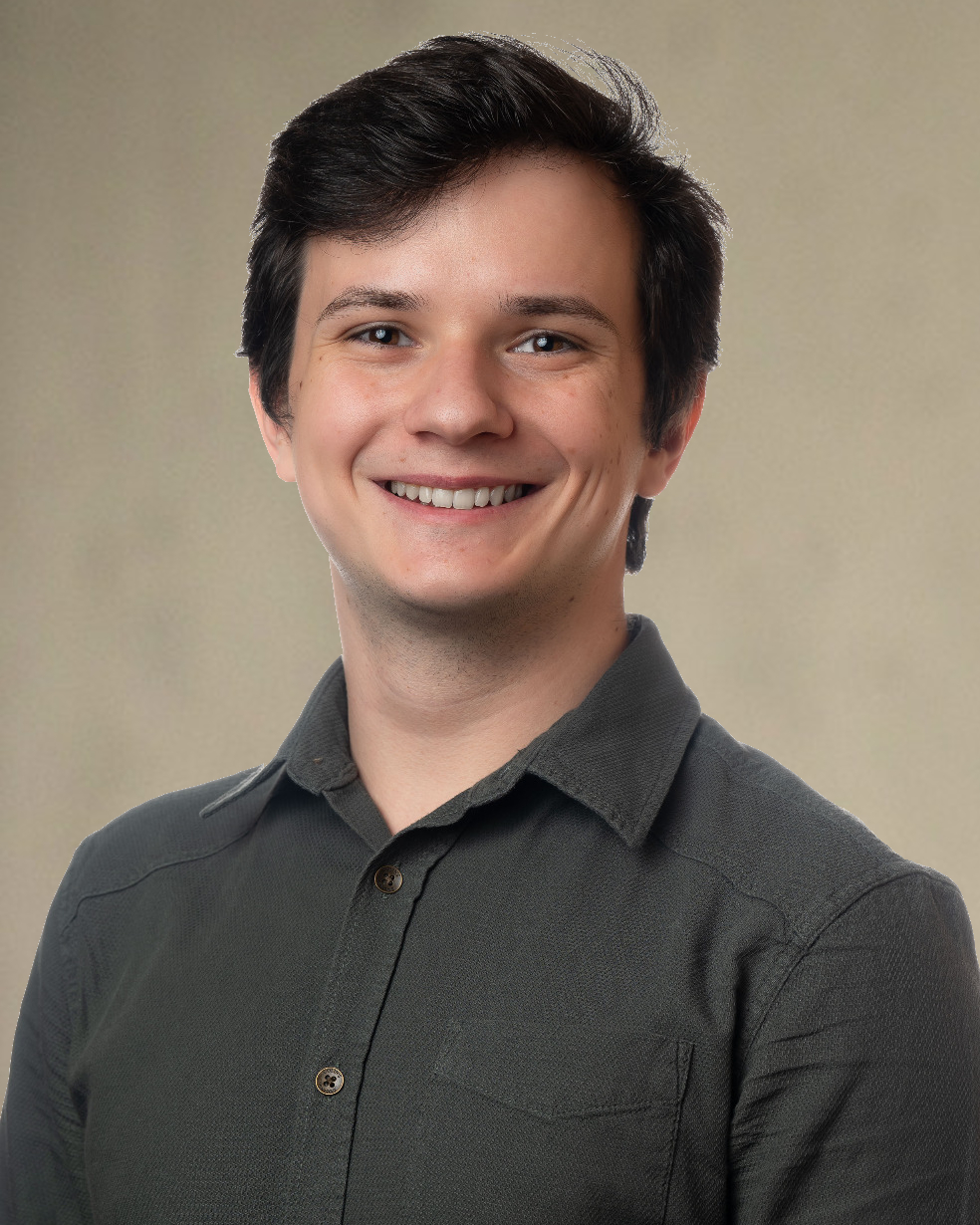}}]
    {\textsc{Mihnea Toader}}~received the B.Sc. degree in computer science and engineering from Technical University Delft with distinction in 2023, and the M.Sc. degree in robotics, systems and control from Swiss Federal Institute of Technology (ETH Zürich) in 2026.  He currently conducts research on robotic perception and navigation systems within the Robotic Systems Lab (RSL) at ETH Zürich.
\end{IEEEbiography}

\begin{IEEEbiography}[{\includegraphics[width=1in,height=1.25in,clip,keepaspectratio]{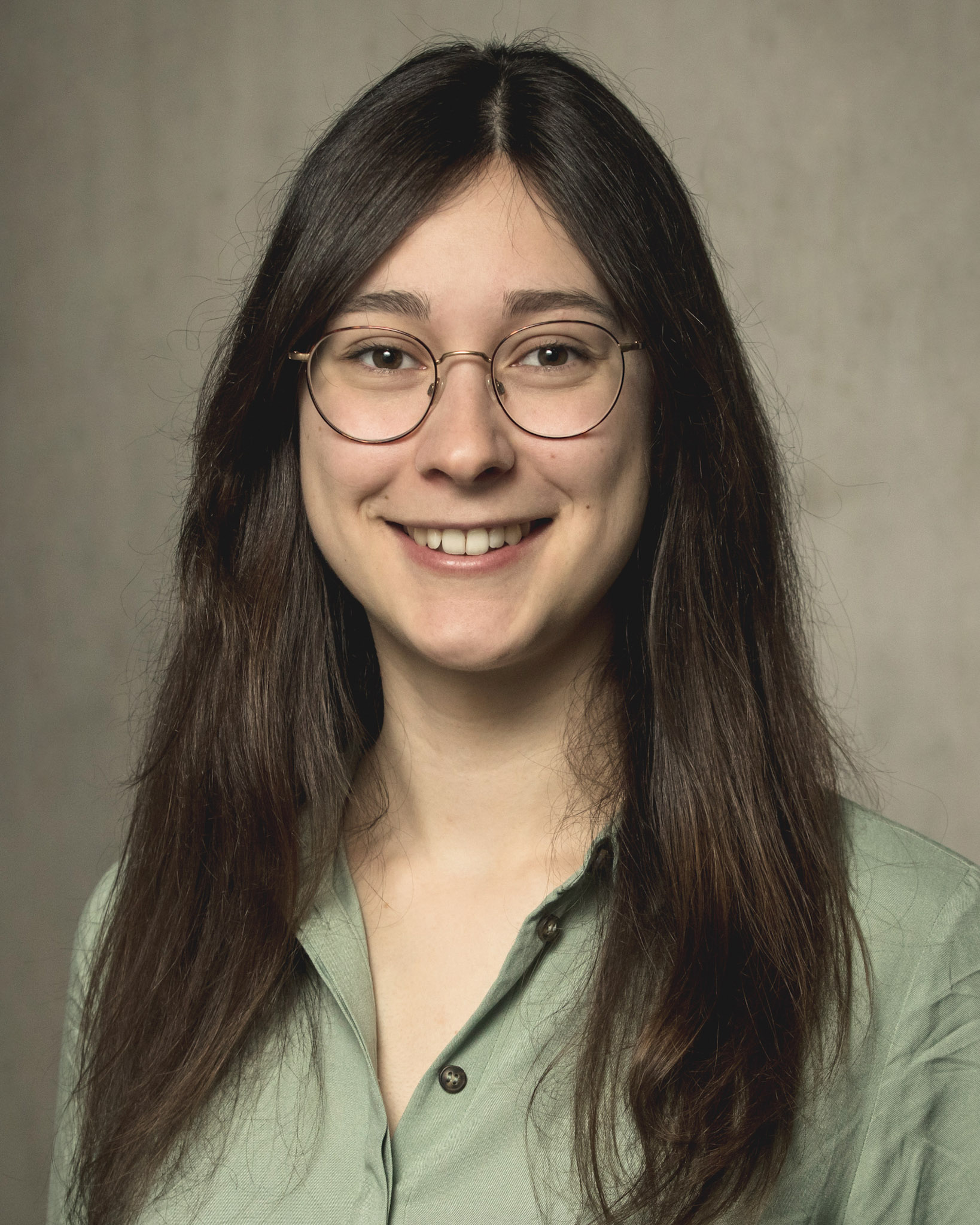}}]
    {\textsc{Julia Richter}}~(Graduate Student Member, IEEE)~received the Dipl.-Ing. degree in mechatronics from the Dresden University of Technology (TUD), Dresden, Germany, in 2023, and is currently pursuing the Ph.D. degree at the Robotic Systems Lab (RSL) at the Swiss Federal Institute of Technology (ETH Zürich), Zürich, Switzerland.

    Her research interests include autonomous navigation in unstructured environments, global path planning under uncertainty, and the integration of local perception with global map reasoning for long-range robotic mobility.
\end{IEEEbiography}

\begin{IEEEbiography}[{\includegraphics[width=1in,height=1.25in,trim=140bp 20bp 141bp 28bp,clip,keepaspectratio]{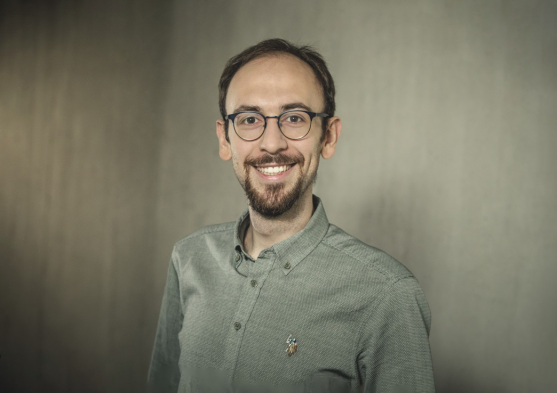}}]
    {\textsc{Turcan Tuna}}~(Graduate Student Member, IEEE)~received the M.Sc. degree in robotics, systems and control from ETH Zürich, Zürich, Switzerland, in 2022, and is currently pursuing the Ph.D. degree at the Robotic Systems Lab (RSL), ETH Zürich. Previously, he earned B.Sc. degrees in mechanical engineering and control and automation engineering at Istanbul Technical University, graduating with distinction in both programs. His research focuses on robust localization, perception, and mapping for mobile robots in challenging environments.
\end{IEEEbiography}

\begin{IEEEbiography}[{\includegraphics[width=1in,height=1.25in,trim=410bp 71bp 351bp 54bp,clip,keepaspectratio]{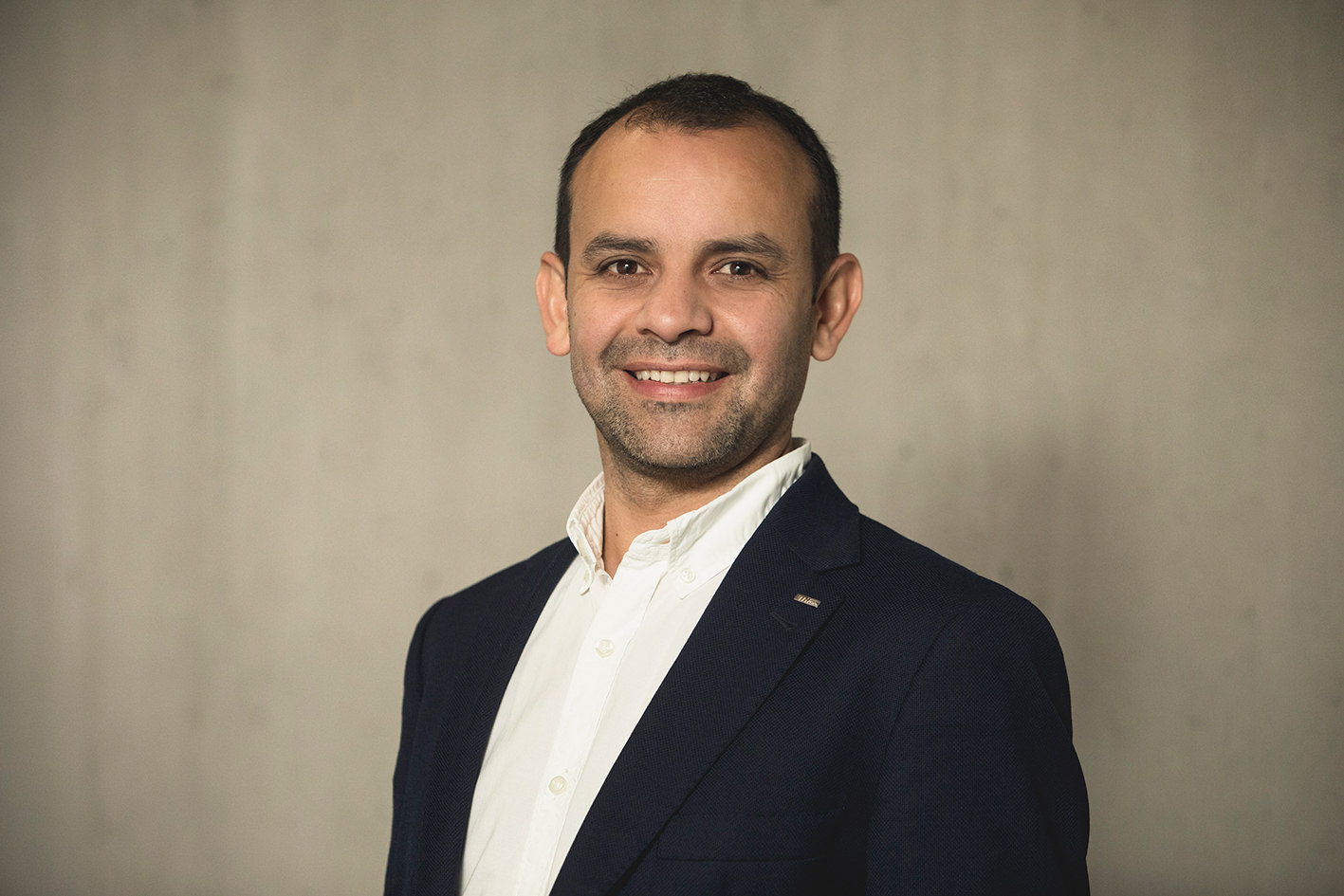}}]
    {\textsc{Cesar Cadena}}~(Member, IEEE)~received his PhD (2011) in computer science from the University of Zaragoza, Spain. He is a Senior Scientist at ETH Zurich. His research focuses on equipping robots to understand and interact with an ever-changing, open world. Building on his work on simultaneous localization and mapping, place recognition, and semantic and persistent mapping, he develops robotic scene understanding that goes beyond geometry and semantics to the physical properties that matter for action. His current interests include self-supervised learning of terrain properties from the robot's own experience, foundation models for robotics and open-world spatial AI, reliable perception under novel conditions, and physically perceptive policies for legged locomotion and mobile manipulation.
\end{IEEEbiography}

\begin{IEEEbiography}[{\includegraphics[width=1in,height=1.25in,clip,keepaspectratio]{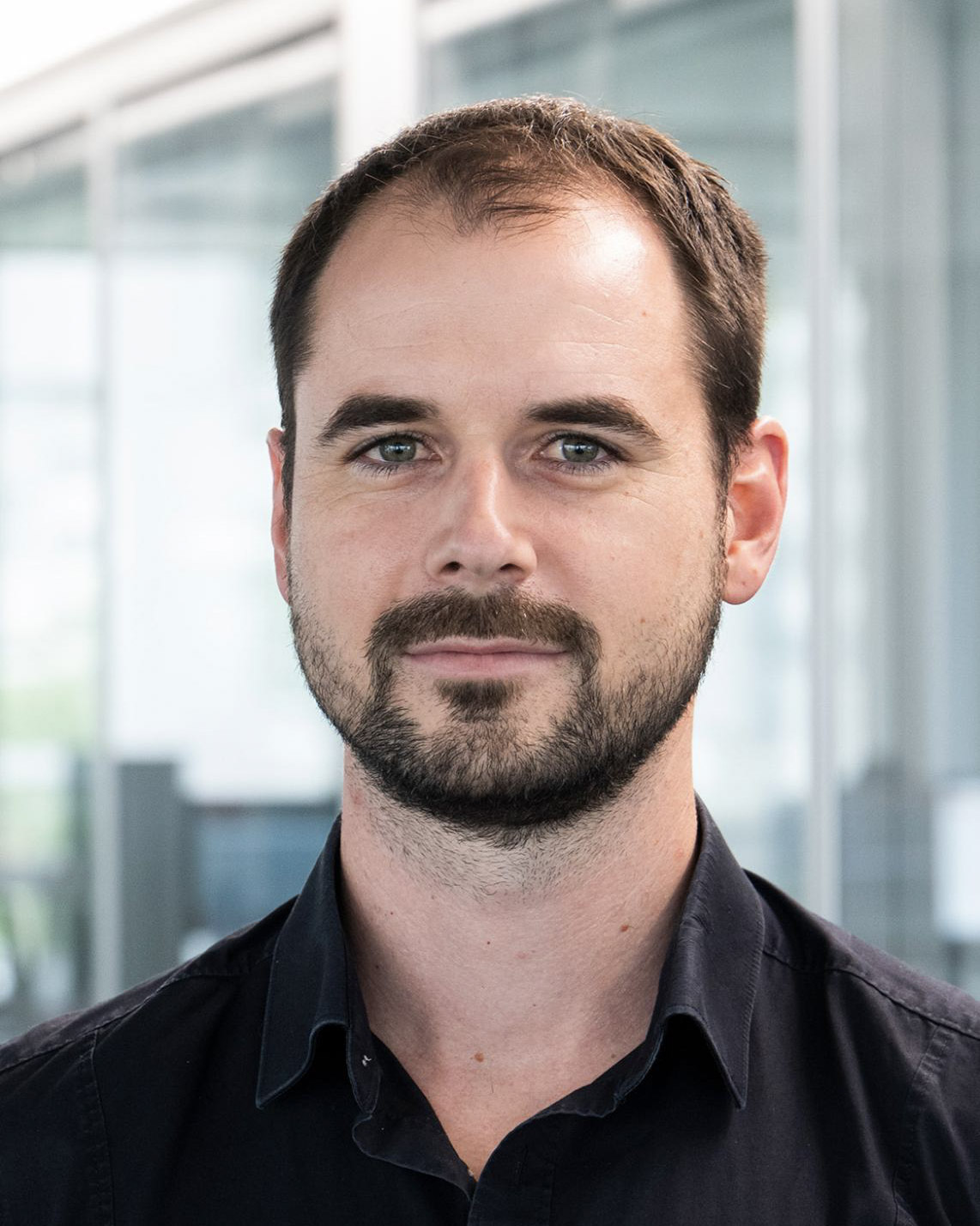}}]
    {\textsc{Marco Hutter}}~(Member, IEEE)~received the M.Sc. and Ph.D. degrees in design, actuation, and control of legged robots from the Swiss Federal Institute of Technology (ETH Zürich), Zürich, Switzerland, in 2009 and 2013, respectively.
    
    He is currently a Professor of robotic systems and the Director of the Center for Robotics, ETH Zürich. He is a Co-Founder of several ETH startups, such as ANYbotics in Zurich, Switzerland, and Gravis Robotics in Zurich. He is also the Director of Boston Dynamics AI Institute Zurich Office, Zurich. He is the Principal Investigator of the NCCRs robotics, automation, and digital fabrication. His research interests include the development of novel machines and actuation concepts, along with the underlying control, planning, and machine learning algorithms for locomotion and manipulation. Dr. Hutter was a recipient of the ERC Starting Grant and the winner of the DARPA SubT Challenge.
\end{IEEEbiography}

\fi

\end{document}